\documentclass{article}

\usepackage{microtype}
\usepackage{graphicx}
\usepackage{subfigure}
\usepackage{booktabs} 

\usepackage{hyperref}

\usepackage[accepted]{icml2024}

\usepackage{amsmath}
\usepackage{amssymb}
\usepackage{mathtools}
\usepackage{amsthm}
\usepackage{mathrsfs}
\usepackage{bm} 
\newcommand{\vct}[1]{\boldsymbol{#1}} 
\newcommand*\infm{\mathrm{d}}
\newcommand{\blue}[1]{\textcolor{blue}{#1}}
\newcommand{\orange}[1]{\textcolor{orange}{#1}}
\definecolor{darkred}{RGB}{162, 0, 0}
\newcommand{\darkred}[1]{\textcolor{darkred}{#1}}
\usepackage{multicol}
\usepackage{lipsum}
\usepackage{stfloats}
\usepackage[small,bf]{caption}
\usepackage{enumitem}
\usepackage{multirow}

\makeatletter
\newcommand*\bigcdot{\mathpalette\bigcdot@{.5}}
\newcommand*\bigcdot@[2]{\mathbin{\vcenter{\hbox{\scalebox{#2}{$\m@th#1\bullet$}}}}}
\makeatother

\usepackage[capitalize,noabbrev]{cleveref}

\theoremstyle{plain}

\theoremstyle{definition}

\theoremstyle{remark}

\allowdisplaybreaks

\usepackage[textsize=tiny]{todonotes}

\usepackage[normalem]{ulem}

\icmltitlerunning{Correlational Lagrangian Schr\"odinger Bridge: 
Learning Dynamics with Population-Level Regularization}

\begin{document}

\twocolumn[
\icmltitle{Correlational Lagrangian Schr\"odinger Bridge:\\
Learning Dynamics with Population-Level Regularization}



\icmlsetsymbol{equal}{*}

\begin{icmlauthorlist}
\icmlauthor{Yuning You}{tamu}
\icmlauthor{Ruida Zhou}{ucla}
\icmlauthor{Yang Shen}{tamu,tamu_cse}
\end{icmlauthorlist}

\icmlaffiliation{tamu}{Department of Electrical and Computer Engineering, Texas A\&M University}
\icmlaffiliation{tamu_cse}{Department of Computer Science and Engineering, Texas A\&M University}
\icmlaffiliation{ucla}{Department of Electrical and Computer Engineering, University of California, Los Angeles}

\icmlcorrespondingauthor{Yang Shen}{yshen@tamu.edu}


\vskip 0.3in
]



\printAffiliationsAndNotice{}  

\begin{abstract}
Accurate modeling of system dynamics holds intriguing potential in broad scientific fields including cytodynamics and fluid mechanics.
This task often presents significant challenges when
(i) observations are limited to \textit{cross-sectional} samples (where individual trajectories are inaccessible for learning),
and moreover,
(ii) the behaviors of individual particles are \textit{heterogeneous} (especially in biological systems due to biodiversity).
To address them, we introduce a novel framework dubbed \textbf{correlational Lagrangian Schr\"odinger bridge} (\textbf{CLSB}), aiming to seek for the evolution ``bridging" among cross-sectional observations, while regularized for the minimal \textit{population} ``cost".
In contrast to prior methods relying on \textit{individual}-level regularizers for all particles \textit{homogeneously} (e.g. restraining individual motions),
CLSB operates at the population level admitting the heterogeneity nature, resulting in a more generalizable modeling in practice.
To this end, our contributions include 
\textbf{(1)} a new class of population regularizers capturing the temporal variations in multivariate relations, with the tractable formulation derived,
\textbf{(2)} three domain-informed instantiations based on genetic co-expression stability, and \textbf{(3)} an integration of population regularizers into data-driven generative models as constrained optimization, and a numerical solution, with further extension to conditional generative models.
Empirically, we demonstrate the superiority of CLSB in single-cell sequencing data analyses such as simulating cell development over time and predicting cellular responses to drugs of varied doses.
\end{abstract}

\begin{figure}[t] 
    \centering 
    \includegraphics[width=1\linewidth]{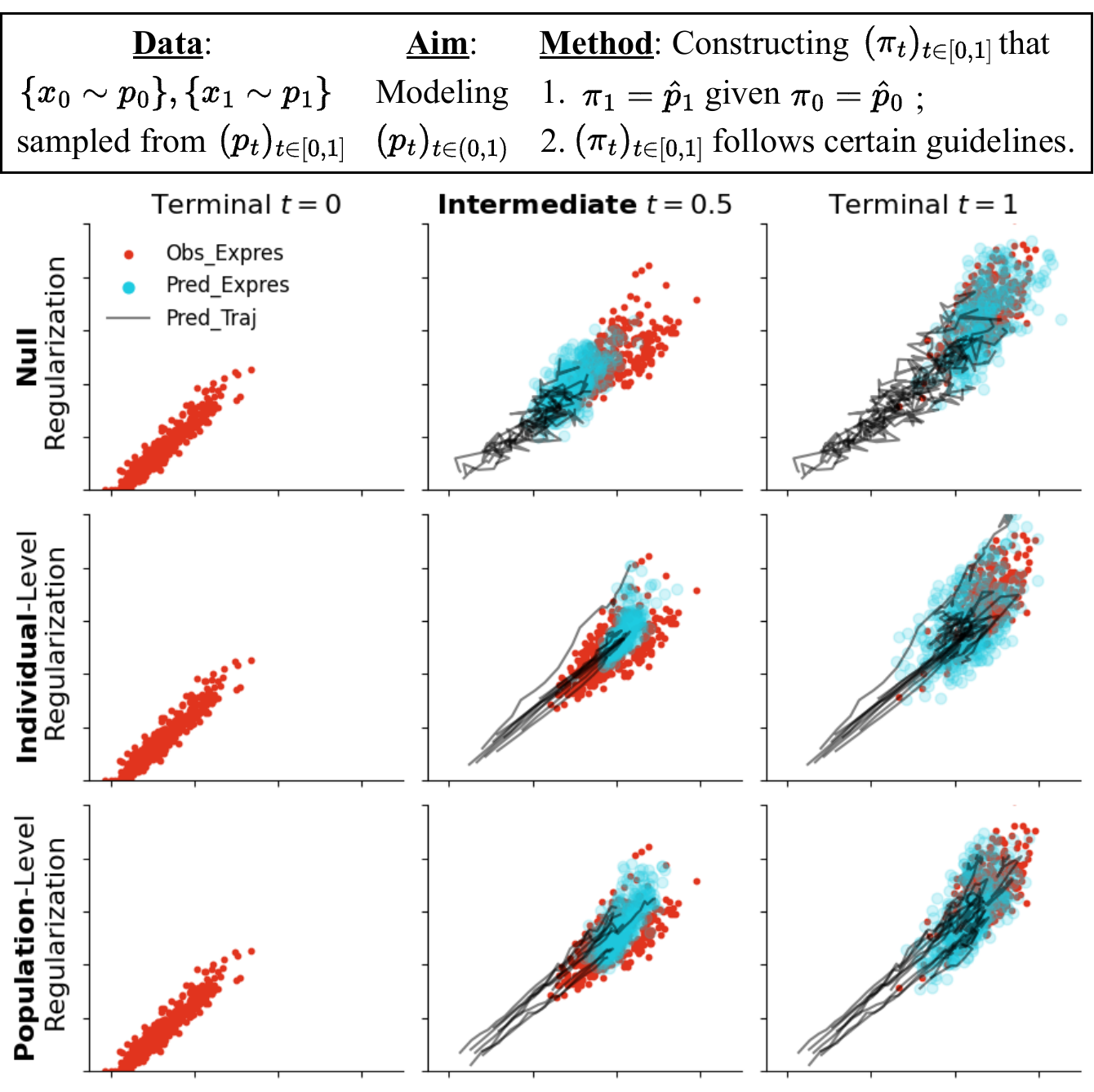}
    \vspace{-1.5em}
    \caption{In-silico simulation of the expressions of gene ABCA3 (x-axis) and A1BG (y-axis) during the embryonic stem cell development with different regularizations.
    Notably, our proposed population-level regularization facilitates a more accurate estimation of distributions, with quantitative evidence detailed in Sec. \ref{sec:experiments}.}
    \label{fig:teaser}
    \vspace{-0.5em}
\end{figure}

\section{Introduction}
Understanding the time evolution of systems of particles plays a fundamental role in numerous scenarios of scientific discovery, e.g., cytodynamics \cite{la2018rna} and fluid mechanics \cite{kundu2015fluid}.
In this paper, the focal problem is to model the dynamic processes of systems based on population samples observed at certain timestamps, i.e., the cross-sectional samples \cite{tong2020trajectorynet,koshizuka2022neural} (see Fig. \ref{fig:teaser} and Sec. \ref{sec:preliminaries} for more details).
This is exemplified in unraveling cellular states for biological and therapeutic insights \cite{keller2005embryonic,till1980hemopoietic}, for which current single-cell sequencing data can only access expression for cell populations rather than individual cell  dynamics \cite{macosko2015highly,chen2015spatially}.

In lack of the ground-truth trajectories for direct supervision, existing approaches attempt to ``bridge" among cross-sectional observations guided by certain principled regularizers (determined by domain priors).
Typical principles include the restraint of motions \cite{schiebinger2019optimal,yang2018scalable}, and the alignment to empirical densities or velocities \cite{tong2020trajectorynet,koshizuka2022neural},
which is formulated at the \textit{microscopic} level, i.e. for states of individual particles, and then applied \textit{homogeneously} to all.
This however could be less rational where the \textit{heterogeneity} of individuals is present (especially for biological systems).  
For instance, in cell clusters, the behaviors of cells could be very distinct from each other in nature due to biodiversity, yet are often modeled to be similar \cite{gaston2013biodiversity,purvis2000getting}.

In view of such a gap, we hypothesize that the principled regularizer, if appropriately formulated at the \textit{macroscopic} level (v.s. microscopic), i.e. for states of population (v.s. individuals), can lead to more generalizable modeling of dynamics in heterogeneous systems.
The logic of the hypothesis is directly related to the needs:
As the ensembles of individuals, population states (i) admit both the necessary domain priors and heterogeneity of individual particles,  
but also importantly, (ii) are capable to accommodate a wider range of domain-specific knowledge evident at the macroscopic level, e.g., the co-expression relations among genes of cellular systems, derived from 
bulk sequencing techniques \cite{stuart2003gene,horvath2008geometric}.
We therefore are inspired to contribute as follows.

\textbf{Contributions.}
We propose a novel learning framework dubbed \textbf{correlational Lagrangian Schr\"odinger bridge} (\textbf{CLSB}) to model dynamics of systems, by \textit{fitting into cross-sectional data while regularized with principled guidance at the population level}, informally as (formally as Opt. \eqref{eq:correlational_lagrangian_schrodinger_bridge}):
\begin{align*}
    & \min_{(\pi_t)_{t\in[0,1]}} \int \overbrace{\Big| \frac{\infm}{\infm t} \underbrace{\mathbb{E}_{\pi_t}[h(\vct{\mathrm{x}})]}_{\text{Population states}} \Big|^2}^{\text{Population guidance}} \infm t, \text{ s.t. } \overbrace{\pi_0 = \hat{p}_0, \pi_1 = \hat{p}_1}^{\text{Data fitting}}.
\end{align*}
To the best of our knowledge, CLSB is the \textit{first} framework to incorporate population-level guidance into dynamic modeling, with \textit{substantial} benefits demonstrated for heterogeneous systems in practice.
In proposing CLSB, we tackle the technical challenges of (1) the tractable formulation of population regularizers, (2) domain-informed instantiations, and (3) integration for data-driven generative modeling.

\textbf{(1)}
\textit{How to \underline{formulate} the principled regularizer at the population level?}
Motivated by the principle of least action \cite{schiebinger2019optimal,yang2018scalable}, we propose to \textit{conserve population states} when bridging across cross-sectional data,
where the conservation is measured by temporal variations of certain statistical characteristics.
Accordingly, we introduce a new category of population regularizers termed \textit{correlational Lagrangian}, which is designed to capture the extent of temporal changes in multivariate relations expressed as moments \cite{parzen1999stochastic,kumar2015stochastic} (Sec. \ref{sec:principle_of_least_population_action}).
For the tractable computation of correlational Lagrangian, we further derive the \textit{analytical expression} under mild assumptions (Sec. \ref{sec:correlational_lagrangian_schrodinger_bridge}).

(\textbf{2})
\textit{How to \underline{instantiate} the population regularizer with domain-informed priors?}
The generic formulation of the correlational Lagrangian is highly versatile, that is able to characterize arbitrary multivariate relations in arbitrary orders.
Inspired by the concept of \textit{co-expression stability} in genetics that the co-expression relations among genes could be robust to environments
\cite{patil2011assessing,srihari2012temporal}, we propose to enforce conservation on the states of covariance, by focusing on the 1st- and 2nd-order variations of bivariate statistics, termed \textit{covariance kinetics} (Sec. \ref{sec:instantiations_and_biological_insights}).
We also leverage the existing evidence of co-expression by constructing \textit{covariance potential}, which enforces alignment between the modeled covariance and the observed interactions from literature (Sec. \ref{sec:covariance_potential}).

(\textbf{3})
\textit{How to \underline{integrate} population guidance into data-driven generative modeling of dynamics?}
Following \cite{koshizuka2022neural}, we formulate a \textit{constrained optimization} problem for integration referred as CLSB, which is designed to minimize correlational Lagrangian subject to the constraints imposed by cross-sectional data,
where the dynamics are parametrized using neural stochastic differential equations (SDEs) \cite{li2020scalable,tzen2019neural}.
To solve CLSB, we propose a numerical \textit{approximation} via unconstrained optimization (Sec. \ref{sec:unconstrained_optimization_with_regularization}).

Furthermore, we extend CLSB for \underline{conditional} generation, by re-engineering neural SDEs for the additional input of conditions. Empirically, we validate that CLSB outperforms state-of-the-art competitors in the experiments of (unconditional) developmental simulation and (conditional) drug-response prediction for cellular systems (Sec. \ref{sec:experiments}).

\section{Preliminaries}
\label{sec:preliminaries}

\textbf{Notations.}
Let the upright letters ($\mathrm{x}$) 
denote  random variables and the italicized ($x$)
denote their realizations.  The boldfaced letters denote vectors if in lowercase ($\vct{x}$) or matrices if in uppercase ($\vct{X}$), as opposed to the regular ($x$) for scalars.  
We use superscripts with brackets ($\vct{x}^{(i)}$) to differentiate multiple realizations of a random vector  $\vct{\mathrm{x}}$, and subscripts with square brackets ($x_{[i]}$ or $x_{[i,j]}$) to identify indexed elements in vector $\vct{x}$ or matrix $\vct{X}$. 
$\nabla$ denotes the divergence operator  $\nabla = [..., \frac{\partial}{\partial x_{[i]}}, ...]^\top$, $\cdot$ denotes vector inner product, and $\bigcdot$ denotes matrix inner product.

\textbf{Data generation from dynamics.}
Let's assume that data are generated from a stochastic process $(\vct{\mathrm{x}}_t)_{t \in [0, 1]}$ following the distribution $(p_t)_{t \in [0, 1]}$
and obeying the dynamics below:
\begin{equation} \label{eq:sde}
    \infm \vct{\mathrm{x}}_t = \vct{f}_t(\vct{\mathrm{x}}_t) \infm t + \vct{G}_t (\vct{\mathrm{x}}_t) \infm \vct{\mathrm{w}}_t,
\end{equation}
where $\vct{\mathrm{x}}_t \in \mathbb{R}^d$, $\vct{f}_t: \mathbb{R}^d \rightarrow \mathbb{R}^d$ is the drift function, $(\vct{\mathrm{w}}_t)_{t \in [0, 1]}$ is a Wiener process in $\mathbb{R}^{d_\text{wie}}$, and $\vct{G}_t: \mathbb{R}^d \rightarrow \mathbb{R}^{d \times d_\text{wie}}$ is the diffusion function.
Consequently the evolution of marginal distribution $p_t$ satisfies the Fokker--Planck equation \cite{risken1996fokker}:
\begin{align} \label{eq:fokker_planck}
    \frac{\partial}{\partial t}p_t(\vct{x}) = & - \nabla \cdot ( p_t(\vct{x}) \vct{f}_t(\vct{x}) ) \notag \\ 
    & + \frac{1}{2} (\nabla \nabla^\top) \bigcdot ( p_t(\vct{x}) \vct{G}_t(\vct{x}) \vct{G}_t^\top(\vct{x}) ).
\end{align}
\textbf{Generative modeling via Schr\"odinger bridge.}
As stated in Eq. \eqref{eq:fokker_planck}, the distribution $(p_t)_{t \in [0, 1]}$ is characterized by the drift and diffusion terms in Eq. \eqref{eq:sde}.
Thus, by parametrizing $\vct{f}_t(\cdot)$ and $\vct{G}_t(\cdot)$
with neural networks $\vct{v}_t(\cdot ; \theta)$ and $ \vct{\Sigma}_t(\cdot ; \theta)$, respectively, and
with observations from the finite-dimensional distribution as 
\begin{align*}
    \mathcal{D}_\text{fdim} = \{ \vct{x}_t^{(i)}: \; & t \in \{t_1, ..., t_s\}, i \in \{1, ..., n\}, \\
    & (\vct{x}_{t_1}^{(i)}, ..., \vct{x}_{t_s}^{(i)}) \sim p_{t_1, ..., t_s} \},
\end{align*}
where $t_1=0, t_s=1$, a line of prior works attempt to construct the generative model $(\pi_t)_{t \in [0,1]}$ 
via solving the collective form of the (static) Schr\"odinger bridge problem as \cite{de2021diffusion,liu2022deep}:
\begin{align} \label{eq:schrodinger_bridge}
    \min_\theta \; & \frac{1}{s-1} \sum_{i=1}^{s-1} \mathrm{KL} ( \pi_{t_i, t_{i+1}} || \hat{p}_{t_i, t_{i+1}} ), \notag \\
    \text{s.t.} \quad & \pi_t(\cdot), \vct{v}_t(\cdot ; \theta), \vct{\Sigma}_t(\cdot ; \theta) \text{ satisfy Eq. \eqref{eq:fokker_planck}}, \pi_t \in \mathscr{P}(\mathbb{R}^d), \notag \\
    & t \in [0,1], \quad\quad \pi_{t_i} = \hat{p}_{t_i}, i \in \{1, ..., s\},
\end{align}
where $\hat{p}_{t_1, ..., t_s}$ is the empirical distribution of $\mathcal{D}_\text{fdim}$, and $\mathscr{P}(\mathbb{R}^d)$ is the probability simplex in $\mathbb{R}^d$.
Conceptually, Opt. \eqref{eq:schrodinger_bridge} requires $(\pi_t)_{t \in [0,1]}$ to align with the reference joint distribution $\hat{p}_{t_i, t_{i+1}}$ as well as marginal $\hat{p}_{t_i}$ of the data.

\textbf{Lagrangian Schr\"odinger bridge for cross-sectional data.}
Trajectory observations $\mathcal{D}_\text{fdim}$ from the finite-dimensional distribution are not always available.
In practice, data might be only observed from the marginal distributions
\begin{equation*}
    \mathcal{D}_\text{marg} = \{ \vct{x}_t^{(i)}:  t \in \{t_1, ..., t_s\}, i \in \{1, ..., n\}, \vct{x}_t^{(i)} \sim p_t \},
\end{equation*}
For such cross-sectional observations, Opt.~\eqref{eq:schrodinger_bridge} is not applicable since the reference distributions $\hat{p}_{t_i, t_{i+1}}, i \in \{1, ..., s-1\}$ in the objective are not available.
Accordingly,
existing solutions propose to solve an alternative optimization problem called  Lagrangian Schr\"odinger bridge (LSB), which adopts the \textit{principled regularizer of least action} instead 
to guide the generative modeling of dynamics as \cite{koshizuka2022neural,neklyudov2023action}:
\begin{align} \label{eq:lagrangian_schrodinger_bridge}
    \min_\theta \; & \frac{1}{(s-1)d} \sum_{i=1}^{s-1} \sum_{j=1}^d \int_{t_i}^{t_{i+1}} L_\text{ind}(\pi_t, j, m) \infm t, \notag \\
    \text{where} \quad & L_\text{ind}(\pi_t, j, m) = \overbrace{\orange{\mathbb{E}_{\pi_t}}}^{\text{Summarization}} \orange{[} \overbrace{ \blue{\Big| \frac{\infm}{\infm t}} ((\mathrm{x}_{t, [j]})^m ) \blue{\Big|^2}}^{\substack{\text{Lagrangian to measure} \\ \text{individual action}}} \orange{]},   \notag \\
    \text{s.t.} \quad & \text{Constraints in Opt. \eqref{eq:schrodinger_bridge}},
\end{align}
where $\hat{p}_{t_1, ..., t_s}$ is overwrote as the empirical distribution of $\mathcal{D}_\text{marg}$.
It is typical to set $m=1$ and approximate the Lagrangian with the expected value as $| \frac{\infm}{\infm t} \mathrm{x}_{t, [j]} |^2 \approx | v_{t, [j]}(\vct{\mathrm{x}}_t ; \theta) |^2 + \vct{\Sigma}_{t, [j,:]}^\top(\vct{\mathrm{x}}_t ; \theta) \vct{\Sigma}_{t, [j,:]}(\vct{\mathrm{x}}_t ; \theta)$ to restrain individual motions \cite{mikami2008optimal,tong2020trajectorynet}.
More related works are detailed in Append. \ref{append:related_works}.

\section{Methods}
\subsection{The Principle of Least Population Action}
\label{sec:principle_of_least_population_action}
The LSB problem \eqref{eq:lagrangian_schrodinger_bridge} enforces the least actions for individual particles during the evolution, i.e., the conservation of individual states.
Two assumptions could be violated in real-world applications (e.g. single-cell sequencing data).
First, effective principles at the individual level may be highly limited in availability, e.g., measuring the behavior of individual cells within clusters can prove challenging.
Second, particles could be heterogeneous in nature \cite{gaston2013biodiversity,purvis2000getting}, while
the simple regularizer $L_\text{ind}(\cdot)$ is formulated for individual states (via action measurement $|\frac{\infm}{\infm t} (\cdot)|^2$) and then applied homogeneously to all particles (via summarization $\mathbb{E}_{\pi_t}[\cdot]$).


To address the lack of effective individual-level principles and to accommodate the heterogeneity among individuals, we propose to shift the focus of guidance to population-level and reorient the emphasis of conservation strategies to population states.
Specifically, we formulate an optimization problem with population regularizer $L_\text{pop}(\cdot)$, by interchanging the order of action measurement $|\frac{\infm}{\infm t} (\cdot)|^2$ and summarization $\mathbb{E}_{\pi_t}[\cdot]$ in $L_\text{ind}(\cdot)$ as follows:
\begin{align} \label{eq:population_lagrangian_schrodinger_bridge}
    \min_\theta \; & \frac{1}{(s-1)d} \sum_{i=1}^{s-1} \sum_{j=1}^d \int_{t_i}^{t_{i+1}} L_\text{pop}(\pi_t, j, m) \infm t, \notag \\
    \text{where} \quad & L_\text{pop}(\pi_t, j, m) = \underbrace{ \blue{\Big| \frac{\infm}{\infm t}} \overbrace{\orange{\mathbb{E}_{\pi_t} [} ( \mathrm{x}_{t, [j]} )^m \orange{]}}^{\text{Population state}} \blue{\Big|^2}}_{\text{Action measurement}}, \notag \\
    \text{s.t.} \quad & \text{Constraints in Opt. \eqref{eq:schrodinger_bridge}},
\end{align} 
where it is assumed $\int | x_{[j]} |^m \pi_t(\vct{x}) d\vct{x} < \infty, j \in \{1, ..., d\}$ \cite{spanos2019probability} for the interchangeability.
The proposed population-level regularizer $L_\text{pop}(\cdot)$ essentially captures the temporal variations in certain population characteristics, contrasting with the focus on individual-state dynamics in $L_\text{ind}(\cdot)$.
In this context, the population state is quantified by the $m$th-order moment of each variable $j$, the determinacy of which is extensively studied in the Hamburger moment problem \cite{shohat1950problem,akhiezer2020classical}.
Thus, Opt. \eqref{eq:population_lagrangian_schrodinger_bridge} aims to find the evolution $(\pi_t)_{t \in [t_i,t_{i+1}]}$ between terminal distributions $\pi_{t_i} = \hat{p}_{t_i}$ and $\pi_{t_{i+1}} = \hat{p}_{t_{i+1}}$ such that the characteristics of distributions evolve smoothly.

Moving beyond Opt. \eqref{eq:population_lagrangian_schrodinger_bridge} we will present a conceptually more complete and computationally tractable formulation for population regularizers in the next subsection.

\subsection{Correlational Lagrangian Schr\"odinger Bridge}
\label{sec:correlational_lagrangian_schrodinger_bridge}
\textbf{Conservation of correlation during evolution.}
The initial extension from individual to population guidance in Opt. \eqref{eq:population_lagrangian_schrodinger_bridge} falls short in capturing the relations among data variables, which accounts for a rich family of observed behaviors especially in biological systems \cite{patil2011assessing,srihari2012temporal}.
Thus, we propose to extend the objective formulation in Opt. \eqref{eq:population_lagrangian_schrodinger_bridge} by involving multivariate relations, resulting in the optimization problem referred as correlational Lagrangian Schr\"odinger bridge (CLSB) as:
\begin{align} \label{eq:correlational_lagrangian_schrodinger_bridge}
    \min_\theta \; & \frac{1}{(s-1)|\mathcal{M}|} \sum_{i=1}^{s-1} \sum_{\widetilde{\mathcal{M}} \in \mathcal{M}} \int_{t_i}^{t_{i+1}} L_\text{corr}(\pi_t, \widetilde{\mathcal{M}}, k) \infm t, \notag \\
    \text{where} \quad & \darkred{L_\text{corr}(\pi_t, \widetilde{\mathcal{M}}, k)} = \underbrace{\Big| \frac{\infm^{\darkred{k}}}{\infm t^{\darkred{k}}} \overbrace{\mathbb{E}_{\pi_t} [ \darkred{\prod_{(j, m) \in \widetilde{\mathcal{M}}}} (\mathrm{x}_{t, [j]})^m ]}^{\text{Correlational characteristic}} \Big|^2}_{\text{Correlational Lagrangian}}, \notag \\
    \text{s.t.} \quad & \text{Constraints in Opt. \eqref{eq:schrodinger_bridge}},
\end{align}
where $k \in \mathbb{Z}_>$, $\mathcal{M} = \{..., \widetilde{\mathcal{M}}_i, ...\}$ that each $\widetilde{\mathcal{M}}_i = \{ (j, \bar{m}_i(j)): j \in \bar{\mathcal{M}}_i \subset \{1,...,d\}, \bar{m}_i: \{1,...,d\} \rightarrow \mathbb{Z}_>\}$ is a multiset consisting of variable indices and their corresponding occurrences, identifying the targeted multivariate relation which is quantified by the mixed moment \cite{parzen1999stochastic,kumar2015stochastic}. Consequently, we refer to $L_\text{corr}(\cdot)$ as \textit{correlational Lagrangian} that captures temporal variations in multivariate correlations.

\begin{table*}[!b]
\begin{center}
\begin{tabular}{|c|}
    \hline \\ [-1ex]
    \textbf{Proposition 1. Analytical Expressions of Correlational Lagrangian
    } \\ [-3ex] \\
    \begin{minipage}{1\textwidth}
      \begin{equation} \label{eq:corr_lagr_1}
        \darkred{L_\text{corr}(\pi_t, \widetilde{\mathcal{M}}, 1)} = \Big| \mathbb{E}_{\pi_t} [ \nabla \Big( \prod_{(j, m) \in \widetilde{\mathcal{M}}} (\mathrm{x}_{t, [j]})^m \Big) \cdot \vct{v}_t(\vct{\mathrm{x}}_t) ] + \frac{1}{2} \mathbb{E}_{\pi_t} [ (\nabla \nabla^\top \Big( \prod_{(j, m) \in \widetilde{\mathcal{M}}} (\mathrm{x}_{t, [j]})^m \Big)) \bigcdot (\vct{\Sigma}_t(\vct{\mathrm{x}}_t) \vct{\Sigma}^\top_t(\vct{\mathrm{x}}_t)) ] \Big|^2.
      \end{equation}
    \end{minipage} \\
    \begin{minipage}{1\textwidth}
    We denote the ordered sequence of operators $\widetilde{\mathcal{S}} = \{..., \Upsilon_{<i, j>}, ...\}$ that $\Upsilon \in \{\nabla, \nabla \nabla^\top\}, i \in \mathbb{Z}_>, j \in \mathbb{Z}_>$ such that
    \end{minipage} \\
    \begin{minipage}{1\textwidth}
        \begin{equation*}
            \Upsilon_{<i, j>} (\vct{x}) = \frac{\infm^i}{\infm t^i} \Upsilon( \prod_{(k, m) \in \widetilde{\mathcal{M}}} (x_{[k]})^m ) \# \frac{\infm^j}{\infm t^j} \gamma(\vct{x}), \quad\quad \# = \Big\{ \begin{smallmatrix} \cdot, \text{ if } \Upsilon = \nabla \\
            \bigcdot, \text{ else if } \Upsilon = \nabla \nabla^\top \end{smallmatrix},
            \gamma(\vct{x}) = \Big\{ \begin{smallmatrix} \vct{v}_t(\vct{x}), \text{ if } \Upsilon = \nabla \\
            \vct{\Sigma}_t(\vct{x}) \vct{\Sigma}^\top_t(\vct{x}), \text{ else if } \Upsilon = \nabla \nabla^\top \end{smallmatrix},
        \end{equation*}
    \end{minipage} \\
    \begin{minipage}{1\textwidth}
    and denote the function $\Gamma(\cdot)$ operating on the ordered sequence $\widetilde{\mathcal{S}}$ and $\circ$ as function composition such that
    \end{minipage} \\
    \begin{minipage}{1\textwidth}
        \begin{equation*}
            \Gamma(\widetilde{\mathcal{S}}) = c_{\widetilde{\mathcal{S}}} \mathbb{E}_{\pi_t} [ \circ_{\Upsilon_{<i, j>} \in \widetilde{\mathcal{S}}} \Upsilon_{<i, j>}( \vct{\mathrm{x}}_t ) ], \quad\quad c_{\widetilde{\mathcal{S}}} = 2^{-|\{\Upsilon_{<i, j>}: \Upsilon_{<i, j>} \in \widetilde{\mathcal{S}}, \Upsilon = \nabla \nabla^\top\}|}.
        \end{equation*}
    \end{minipage} \\
    \begin{minipage}{1\textwidth}
    We then have $L_\text{corr}(\pi_t, \widetilde{\mathcal{M}}, 1) = | \Gamma(\{\nabla_{<0, 0>}\}) + \Gamma(\{(\nabla \nabla^\top)_{<0, 0>}\} ) |^2$.
    We further denote $\mathcal{S}_{<k>}$ as the set of $\widetilde{\mathcal{S}}$ used to compute $L_\text{corr}(\pi_t, \widetilde{\mathcal{M}}, k)$, e.g., $\mathcal{S}_{<1>} = \{ \{\nabla_{<0, 0>}\}, \{(\nabla \nabla^\top)_{<0, 0>}\} \}$, and then we have for $k \ge 2$:
    \end{minipage} \\
    \begin{minipage}{1\textwidth}
        \begin{equation} \label{eq:corr_lagr_k}
            \darkred{L_\text{corr}(\pi_t, \widetilde{\mathcal{M}}, k)} = \Big| \sum_{\substack{\widetilde{\mathcal{S}} \in \mathcal{S}_{<k-1>}, \\ \widetilde{\mathcal{S}} = \widetilde{\mathcal{S}}' \cup \{\Upsilon'_{<i', j'>}\} }} \Gamma(\widetilde{\mathcal{S}}' \cup \Upsilon'_{<i'+1, j'>}) + \Gamma(\widetilde{\mathcal{S}}' \cup \Upsilon'_{<i', j'+1>}) + \Gamma(\widetilde{\mathcal{S}}  \cup \nabla_{<0, 0>}) + \Gamma(\widetilde{\mathcal{S}}  \cup (\nabla \nabla^\top)_{<0, 0>}) \Big|^2.
        \end{equation}
    \end{minipage} \\
    [6ex]
    \hline
\end{tabular}
\end{center}
\end{table*} 

The CLSB formulation \eqref{eq:correlational_lagrangian_schrodinger_bridge} is the more general framework, capable of imposing domain priors for 
arbitrary multivariate relations (specified by $\widetilde{\mathcal{M}}$ in $L_\text{corr}(\cdot)$) in arbitrary order (specified by $k$). 
For instance, by setting $k=1$ and $\mathcal{M} = \{\widetilde{\mathcal{M}}_j: \widetilde{\mathcal{M}}_j = \{(j, m)\}, j = 1,...,d\}$, CLSB degenerates to Opt. \eqref{eq:population_lagrangian_schrodinger_bridge} that involves null correlations.

\textbf{Analytical expression of correlational Lagrangian.}
The current formulation of correlational Lagrangian $L_\text{corr}(\cdot)$ in Opt. \eqref{eq:correlational_lagrangian_schrodinger_bridge} is not yet in a tractable form for practical computation 
due to the existence of the $k$-order time derivative
Our following proposition provides the tractable analytical expression under mild assumptions.

\textbf{Proposition 1.}
For $k=1$, correlational Lagrangian admits the analytical expression in Eq. \eqref{eq:corr_lagr_1} (\darkred{see details in the box below}), if for the set of functions $\mathcal{H} =$
$\{ h(\vct{x}) \pi_t(\vct{x}) \vct{v}_t(\vct{x}),$
$\pi_t(\vct{x}) \vct{D}(\vct{x}) \nabla h(\vct{x}),$
$\pi_t(\vct{x}) \nabla^\top\vct{D}_t(\vct{x}) h(\vct{x}),$
$h(\vct{x}) \vct{D}_t(\vct{x}) \nabla \pi_t(\vct{x}) \}$ ($\theta$ is omitted for simplicity) that $h(\vct{x}) = \prod_{(j, m) \in \widetilde{\mathcal{M}}} (\mathrm{x}_{t, [j]})^m$, $\vct{D}_t(\vct{x}) = \vct{\Sigma}_t(\vct{x}) \vct{\Sigma}_t^\top(\vct{x})$, it satisfies:
\begin{itemize}[leftmargin=*] \vspace{-0.7em}
    \item Continuity. $h' \in \mathcal{H}$ is continuously differentiable w.r.t. $\vct{x}$;
    \vspace{-1.5em}
    \item Light tail. The probability density function $\pi_t(\vct{x})$ is characterized by tails that are sufficiently light, such that $\oint_{S_\infty} h'(\vct{x})  \cdot \infm \vct{a} = 0$ for $h' \in \mathcal{H}$, where $\vct{a}$ is the outward pointing unit normal on the boundary at infinity $S_\infty$.
\end{itemize} \vspace{-0.7em}
For $k \ge 2$, correlational Lagrangian admits the analytical expression in Eq. \eqref{eq:corr_lagr_k} (\darkred{see details in the box below}) in an iterative manner, similarly, if certain conditions of continuity and light tail are satisfied (which are listed in Append. \ref{append:proof_for_proposition_1}).

\textit{Proof.} See Append. \ref{append:proof_for_proposition_1}.

Both conditions are moderate and can be easily ensured by appropriately constructing the architectures of the drift and diffusion functions \cite{schulz2018tutorial,song2020score}.
Thereby, Propos. 1 enables the tractable computation of correlational Lagrangian for practical implementation.

\subsection{Domain-Informed Instantiations of Correlational Lagrangian in Biological Systems}
\label{sec:tractable_objective}
\subsubsection{Covariance Kinetics}
\label{sec:instantiations_and_biological_insights}

\textbf{Conserving bivariate relations for co-expression stability.}
Existing literature in genetics indicates the phenomenon of co-expression stability, i.e., the co-expression among genes could be robust to environments.
For instance, \cite{patil2011assessing,srihari2012temporal} identify the interactive behaviors among ``hub" genes in the gene networks, and genes with static proteins translated,
are relatively insensitive to environmental perturbations.
We numerically validate such phenomena in our dataset (see Append. \ref{append:coexpression_stability} for details).

We are therefore inspired to incorporate such domain knowledge into the population regularizer, by narrowing the focus of correlational Lagrangian specifically on bivariate relations, thereby restraining temporal variations of the states of covariance.
We term it as \textit{covariance kinetics}, which mimics the idea of restricting the ``motion" of the population \cite{frost1961kinetics,lifschitz1983physical}, with two specific forms of instantiation as follows. 

\textbf{Instantiation 1: Restraining the ``velocity" of covariance.}
Let $\mathcal{M}_\text{cov} = \Big\{ \{(i, 1), (j, 1)\}: i \in \{1, ..., d\}, j \in \{1, ..., d\} \Big\}$ for all the pairs among $d$ variables. 
The principled objective in Opt. \eqref{eq:correlational_lagrangian_schrodinger_bridge}, which captures the 1st-order temporal variations of covariance, is expressed as:
\begin{align} \label{eq:covariance_velocity}
    & \sum_{\widetilde{\mathcal{M}} \in \mathcal{M}_\text{cov}}  L_\text{corr}(\pi_t, \widetilde{\mathcal{M}}, 1) = \Big\lVert \frac{\infm}{\infm t} \mathbb{E}_{\pi_t}[ \vct{\mathrm{x}}_t \vct{\mathrm{x}}_t^\top ] \Big\rVert^2_\mathsf{F} \notag \\
    = \; & \Big\lVert \mathbb{E}_{\pi_t}[\vct{\mathrm{x}}_t \vct{v}_t(\vct{\mathrm{x}}_t)^\top + \vct{v}_t(\vct{\mathrm{x}}_t) \vct{\mathrm{x}}_t^\top + \frac{1}{2} \vct{\Sigma}_t(\vct{\mathrm{x}}_t) \vct{\Sigma}^\top_t(\vct{\mathrm{x}}_t) ] \Big\rVert^2_\mathsf{F}.
\end{align}
\textbf{Instantiation 2: Restraining the ``acceleration" of covariance.}
The principled objective, capturing the 2nd-order temporal variations of covariance, is expressed as:
\begin{align} \label{eq:covariance_acceleration}
    & \sum_{\widetilde{\mathcal{M}} \in \mathcal{M}_\text{cov}}  L_\text{corr}(\pi_t, \widetilde{\mathcal{M}}, 2) = \Big\lVert \frac{\infm^2}{\infm t^2} \mathbb{E}_{\pi_t}[ \vct{\mathrm{x}}_t \vct{\mathrm{x}}_t^\top ] \Big\rVert^2_\mathsf{F} \notag \\
    = \; & \Big\lVert \overbrace{\mathbb{E}_{\pi_t}\Big[\vct{\mathrm{x}}_t (\frac{\infm}{\infm t}\vct{v}_t(\vct{\mathrm{x}}_t))^\top + (\frac{\infm}{\infm t}\vct{v}_t(\vct{\mathrm{x}}_t)) \vct{\mathrm{x}}_t^\top}^{\text{Collection of } \Gamma(\widetilde{\mathcal{S}}' \cup \Upsilon'_{<i', j'+1>}) \text{ terms in Eq. \eqref{eq:corr_lagr_k}}} \notag \\
    & + \frac{1}{2} \frac{\infm}{\infm t}( \vct{\Sigma}_t(\vct{\mathrm{x}}_t) \vct{\Sigma}^\top_t(\vct{\mathrm{x}}_t) ) \Big] \quad\quad\quad + \quad \overbrace{\vct{0}}^{\Gamma(\widetilde{\mathcal{S}}' \cup \Upsilon'_{<i'+1, j'>})} \notag \\
    & + \overbrace{\mathbb{E}_{\pi_t}\Big[ \vct{\mathrm{x}}_t (\nabla\vct{v}_t(\vct{\mathrm{x}}_t) \vct{v}_t(\vct{\mathrm{x}}_t))^\top + (\nabla\vct{v}_t(\vct{\mathrm{x}}_t) \vct{v}_t(\vct{\mathrm{x}}_t)) \vct{\mathrm{x}}_t^\top}^{\Gamma(\widetilde{\mathcal{S}} \cup \nabla_{<0, 0>})} \notag \\
    & + 2 \vct{v}_t(\vct{\mathrm{x}}_t) \vct{v}_t(\vct{\mathrm{x}}_t)^\top + \frac{1}{2} \nabla(\vct{\Sigma}_t(\vct{\mathrm{x}}_t) \vct{\Sigma}^\top_t(\vct{\mathrm{x}}_t))_{\underline{i_1 i_2 i_3}} \vct{v}_t^{\underline{i_3}}(\vct{\mathrm{x}}_t) \Big] \notag \\
    & + \overbrace{\mathbb{E}_{\pi_t}\Big[ \nabla\vct{v}_t(\vct{\mathrm{x}}_t) \vct{\Sigma}_t(\vct{\mathrm{x}}_t) \vct{\Sigma}^\top_t(\vct{\mathrm{x}}_t)}^{\Gamma(\widetilde{\mathcal{S}} \cup (\nabla \nabla^\top)_{<0, 0>})} \notag \\
    & + \vct{\Sigma}_t(\vct{\mathrm{x}}_t) \vct{\Sigma}^\top_t(\vct{\mathrm{x}}_t) \nabla^\top\vct{v}_t(\vct{\mathrm{x}}_t) \notag \\
    & + \frac{1}{2} \vct{\mathrm{x}}_t ( \nabla \nabla^\top(\vct{v}_t(\vct{\mathrm{x}}_t))_{\underline{i_1 i_2 i_3}} (\vct{\Sigma}_t(\vct{\mathrm{x}}_t) \vct{\Sigma}^\top_t(\vct{\mathrm{x}}_t))^{\underline{i_2 i_3}} )^\top \notag \\
    & + \frac{1}{2} (\nabla \nabla^\top(\vct{v}_t(\vct{\mathrm{x}}_t))_{\underline{i_1 i_2 i_3}} (\vct{\Sigma}_t(\vct{\mathrm{x}}_t) \vct{\Sigma}^\top_t(\vct{\mathrm{x}}_t))^{\underline{i_2 i_3}}) \vct{\mathrm{x}}_t^\top \notag \\
    & + \frac{1}{4} \nabla \nabla^\top(\vct{\Sigma}_t(\vct{\mathrm{x}}_t) \vct{\Sigma}^\top_t(\vct{\mathrm{x}}_t))_{\underline{i_1 i_2 i_3 i_4}} (\vct{\Sigma}_t(\vct{\mathrm{x}}_t) \vct{\Sigma}^\top_t(\vct{\mathrm{x}}_t))^{\underline{i_3 i_4}} \Big] \Big\rVert^2_\mathsf{F},
\end{align}
where we extend the divergence operator $\nabla$ for vector functions and adopt the Einstein notation for tensor operations \cite{barr1991einstein}.
The matrix-form derivations of instantations \eqref{eq:covariance_velocity} \& \eqref{eq:covariance_acceleration} are based on Propos. 1.


\textbf{Standardized covariance kinetics within a linearly projected space.}
The instantiated objectives \eqref{eq:covariance_velocity} \& \eqref{eq:covariance_acceleration} can be re-written for the standardized covariance, which are more robust to encompass co-expression priors (less suffering from the batch effect in sequencing techniques \cite{zhang2019novel,luo2010comparison}),
e.g. for the velocity term \eqref{eq:covariance_velocity}:
\begin{align} \label{eq:standardized_covariance_velocity}
    & \sum_{\widetilde{\mathcal{M}} \in \mathcal{M}_\text{cov}}  L_\text{corr-std}(\pi_t, \widetilde{\mathcal{M}}, 1) \notag \\
    = \; & \Big\lVert \frac{\infm}{\infm t} \Big( \mathbb{E}_{\pi_t}[ \vct{\mathrm{x}}_t \vct{\mathrm{x}}_t^\top ] - \mathbb{E}_{\pi_t}[ \vct{\mathrm{x}}_t] \mathbb{E}_{\pi_t}^\top[ \vct{\mathrm{x}}_t ] \Big) \Big\rVert^2_\mathsf{F} \notag \\
    = \; & \Big\lVert \mathbb{E}_{\pi_t}[\vct{\mathrm{x}}_t \vct{v}_t(\vct{\mathrm{x}}_t)^\top + \vct{v}_t(\vct{\mathrm{x}}_t) \vct{\mathrm{x}}_t^\top + \frac{1}{2} \vct{\Sigma}_t(\vct{\mathrm{x}}_t) \vct{\Sigma}^\top_t(\vct{\mathrm{x}}_t) ] \notag \\
    & - \mathbb{E}_{\pi_t}[ \vct{\mathrm{x}}_t] \mathbb{E}_{\pi_t}^\top[ \vct{v}_t(\vct{\mathrm{x}}_t) ] - \mathbb{E}_{\pi_t}[ \vct{v}_t(\vct{\mathrm{x}}_t) ] \mathbb{E}_{\pi_t}^\top[ \vct{\mathrm{x}}_t] \Big\rVert^2_\mathsf{F}.
\end{align}
Similarly for the acceleration term \eqref{eq:covariance_acceleration}, the standardized formulation is detailed in Append. \ref{append:acceleration_of_standardized_covariance}.

Furthermore, the objective can be re-written for systems characterized by priors residing in a linearly projected space.
This scenario is common in high-dimensional single-cell sequencing data, which often undergoes principal component analysis \cite{wold1987principal} before subsequent processing.
Specifically, when gene expressions are mapped from the principal components as $\vct{x}_\text{gene} = \vct{W} \vct{x} + \vct{b}$, the projected (and standardized)
$k$-th order
covariance kinetics is then expressed of the matrix form as:
\begin{align} \label{eq:projected_space_with_linear_decoder_w}
    & \sum_{\widetilde{\mathcal{M}} \in \mathcal{M}_\text{cov}}  L_\text{corr-std-proj}(\pi_t, \widetilde{\mathcal{M}}, k) \notag \\
    = \; & \Big\lVert \vct{W} \frac{\infm^k}{\infm t^k} \Big( \mathbb{E}_{\pi_t}[ \vct{\mathrm{x}}_t \vct{\mathrm{x}}_t^\top ] - \mathbb{E}_{\pi_t}[ \vct{\mathrm{x}}_t] \mathbb{E}_{\pi_t}^\top[ \vct{\mathrm{x}}_t ] \Big) \vct{W}^\top \Big\rVert^2_\mathsf{F}.
\end{align}

\subsubsection{Covariance Potential}
\label{sec:covariance_potential}
\textbf{Instantiation 3: Aligning covariance with observed bivariate interactions.}
There exists abundant observed evidence of co-expression relations among genes, sourced from numerous experiments which represent these interactions in a statistical context \cite{mering2003string,oughtred2019biogrid}.
To leverage such prior knowledge, specifically denoted as $\vct{Y} \in [0, 1]^{d \times d}$ where $\vct{Y}_{[i, j]}$ denotes the confidence score of genes $i$ and $j$ being co-expressing,
we construct the principled regularizer termed \textit{covariance potential}, which borrows the idea of enforcing alignment with the ``correct position" of the population states as:
\begin{equation} \label{eq:covariance_potential}
    \sum_{\widetilde{\mathcal{M}} \in \mathcal{M}_\text{cov}} L_\text{corr}(\pi_t, \widetilde{\mathcal{M}}, 0) = - U \Big( \mathbb{E}_{\pi_t}[ \vct{\mathrm{x}}_t \vct{\mathrm{x}}_t^\top ], \vct{Y} \Big),
\end{equation}
where $U(\cdot)$ is the designated potential function detailed in Append. \ref{append:form_of_potential_function},
and the notation $L_\text{corr}(\cdot)$ is reused here, as it was previously undefined for $k=0$.

\begin{table*}[!t]
\begin{center}
\caption{Evaluation in the unconditional generation scenario of modeling embryonic stem-cell development. Reported are Wasserstein distances, where lower values are preferable, with standard deviations in parentheses.  The top-2 best performances in each case and those on average ('A.R.' stands for average rank) are highlighted in \darkred{\textbf{red}}.
Methods are evaluated in the scenarios of all-step prediction on $\pi_{t_i | t_0}$ where $\pi_{t_0} = \hat{p}_{t_0}$, and one-step prediction on $\pi_{t_i | t_{i-1}}$ where $\pi_{t_{i-1}} = \hat{p}_{t_{i-1}}$, $i \in \{1, 2, 3\}$.
}
\vspace{-0.5em}
\label{tab:uncondition_generation}
\resizebox{1\textwidth}{!}{
\begin{tabular}{c | c c c | c c c | c c c}
    \toprule
    \multirow{2}{*}{Methods} & \multicolumn{3}{c|}{All-Step Prediction} & \multicolumn{3}{c|}{One-Step Prediction} & \multirow{2}{*}{\textbf{A.R.}} \\
    & $t_1$ & $t_2$ (Most Challenging) & $t_3$ & $t_1$ & $t_2$ & $t_3$ \\
    \midrule
    Random & 1.873$\pm$0.014 & 2.082$\pm$0.011 & 1.867$\pm$0.011 & 1.870$\pm$0.013 & 2.084$\pm$0.010 & 1.868$\pm$0.012 & 9.5 \\
    OT-Flow & 1.921 & 2.421 & 1.542 & 1.921 & 1.151 & 1.438 & 8.5 \\
    OT-Flow+OT & 1.726 & 2.154 & 1.397 & 1.726 & 1.186 & 1.240 & 7.3 \\
    TrajectoryNet & 1.774 & 1.888 & \darkred{\textbf{1.076}} & 1.774 & 1.178 & 1.315 & 6.5 \\
    TrajectoryNet+OT & 1.134 & \darkred{\textbf{1.336}} & \darkred{\textbf{1.008}} & 1.134 & 1.151 & 1.132 & 3.6 \\
    DMSB & 1.593 & 2.591 & 2.058 & -- & -- & -- & 9.6 \\
    NeuralSDE & 1.507$\pm$0.014 & 1.743$\pm$0.031 & 1.586$\pm$0.038 & 1.504$\pm$0.013 & 1.384$\pm$0.016 & 0.962$\pm$0.014 & 6.1 \\
    NLSB(E) & 1.128$\pm$0.007 & 1.432$\pm$0.022 & 1.132$\pm$0.034 & 1.130$\pm$0.007 & \darkred{\textbf{1.099}}$\pm$0.010 & \darkred{\textbf{0.839}}$\pm$0.012 & 2.6 \\
    NLSB(E+D+V) & 1.499$\pm$0.005 & 1.945$\pm$0.006 & 1.619$\pm$0.016 & 1.498$\pm$0.005 & 1.418$\pm$0.009 & 0.966$\pm$0.016 & 6.6 \\
    \midrule
    CLSB($\alpha_\text{ind}>0$) & \darkred{\textbf{1.099}}$\pm$0.019 & 1.419$\pm$0.028 & 1.132$\pm$0.038 & \darkred{\textbf{1.098}}$\pm$0.018 & 1.117$\pm$0.009 & \darkred{\textbf{0.826}}$\pm$0.010 & \darkred{\textbf{2.5}} \\
    CLSB($\alpha_\text{ind}=0$) & \darkred{\textbf{1.074}}$\pm$0.009 & \darkred{\textbf{1.244}}$\pm$0.016 & 1.255$\pm$0.022 & \darkred{\textbf{1.095}}$\pm$0.009 & \darkred{\textbf{1.106}}$\pm$0.014 & 0.842$\pm$0.012 & \darkred{\textbf{2.1}} \\
    \bottomrule
\end{tabular}} \vspace{-1em}
\end{center}
\end{table*}

\subsection{Numerical Solutions of CLSB}
\label{sec:unconstrained_optimization_with_regularization}

\textbf{Approximation via unconstrained optimization.}
The exact solution of CLSB \eqref{eq:correlational_lagrangian_schrodinger_bridge} remains challenging despite that we provide a tractable objective in Sec. \ref{sec:tractable_objective}, due to its non-convex objective and constraints w.r.t. network parameters.
Thus, we propose a practical solution for approximation via grappling with an unconstrained optimization problem as:
\begin{align} \label{eq:clsb_unconstraint}
    \min_\theta \; & \frac{1}{(s-1)} \sum_{i=1}^{s-1} \Big( L_\text{dist}(\pi_{t_i+1}, \hat{p}_{t_i+1}) \notag \\
    & + \alpha_\text{ind} \frac{1}{d} \sum_{j=1}^d \int_{t_i}^{t_{i+1}} L_\text{ind}(\pi_t, j, 1) \infm t \notag \\
    & + \sum_{k=0}^2 \alpha_{\text{corr}, k} \frac{1}{|\mathcal{M}_\text{cov}|} \sum_{\widetilde{\mathcal{M}} \in \mathcal{M}_\text{cov}} \int_{t_i}^{t_{i+1}} L_\text{corr}(\pi_t, \widetilde{\mathcal{M}}, k) \infm t \Big),
\end{align}
where $L_\text{dist}(\cdot)$ is the distribution discrepancy measure, and $\alpha_\text{ind}, \alpha_{\text{corr}, 0}, \alpha_{\text{corr}, 1}, \alpha_{\text{corr}, 2}$ are the weights for different guidance objectives.
Here we also adopt the individual guidance $L_\text{ind}(\cdot)$ and adjust its weight $\alpha_\text{ind}$ through hyperparameter tuning.
Opt. \eqref{eq:clsb_unconstraint} is solved via gradient descent in practice.

\textbf{Extension to conditional generative modeling.}
We further extend CLSB into the conditional generation scenario, where we are tasked to model $(p_t(\cdot | \vct{\mathrm{c}}))_{t \in [0, 1]}$.
The application encompasses modeling cellular systems in response to perturbations $\vct{\mathrm{c}}$ such as drug treatments or more \cite{srivatsan2020massively,dong2023causal}.
Such extension can be achieved by re-engineering the neural SDEs $\vct{v}_t(\cdot ; \theta), \vct{\Sigma}_t(\cdot ; \theta)$ to input additional featurized conditions, which is re-written as $\vct{v}_t(\cdot, \vct{c} ; \theta), \vct{\Sigma}_t(\cdot, \vct{c} ; \theta)$, without altering the rest components in the framework.

\section{Experiments}
\label{sec:experiments}
We evaluate the proposed CLSB \eqref{eq:clsb_unconstraint}
in two real-world applications of modeling cellular systems in the unconditional (Sec. \ref{sec:exp_unconditional_generation}) and conditional generation scenarios (Sec. \ref{sec:exp_conditional_generation}).

\begin{figure*}[t] 
    \centering 
    \includegraphics[width=1\linewidth]{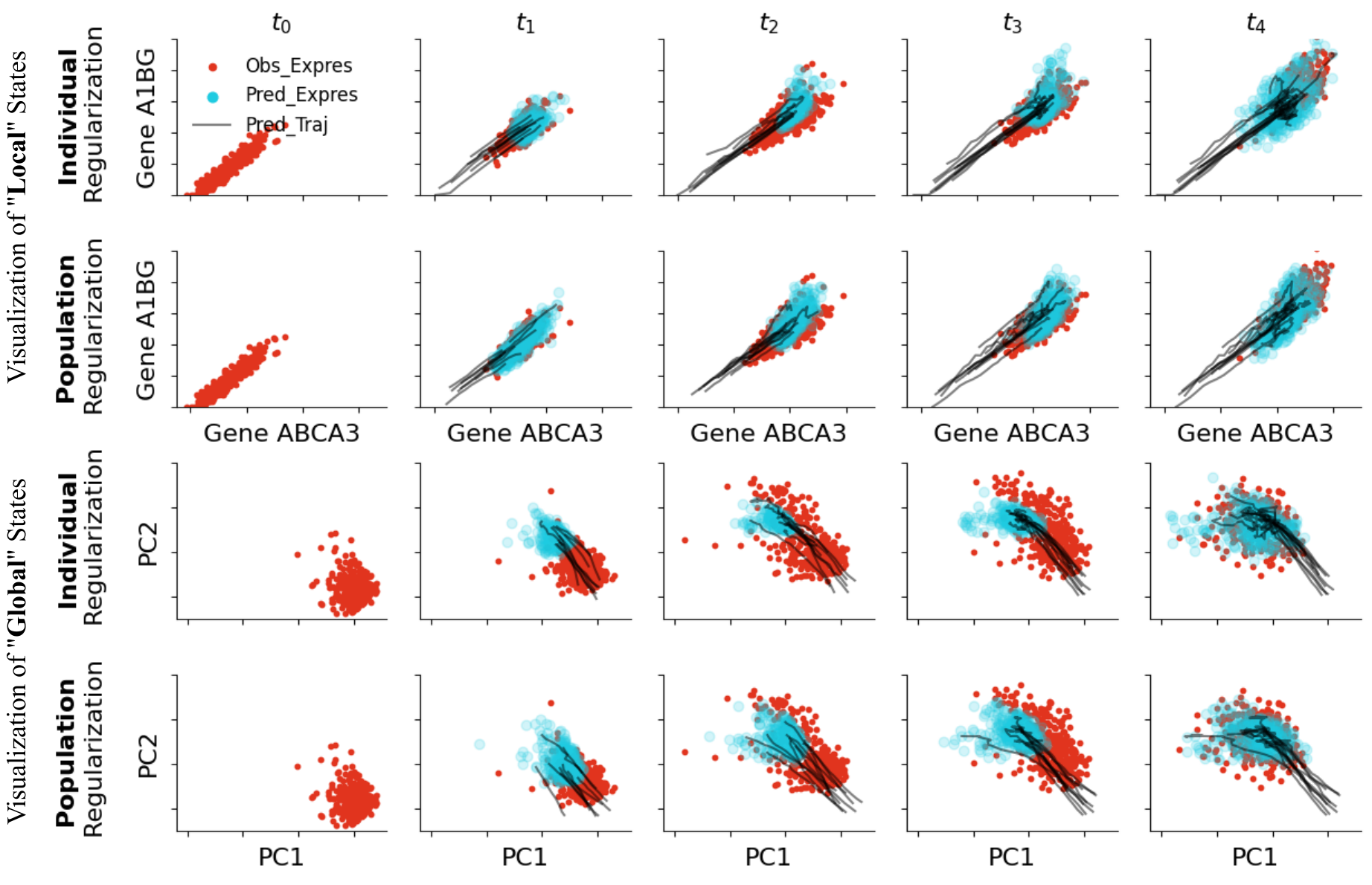}
    \vspace{-1.5em}
    \caption{Visualization of the simulated gene expressions and trajectories with different methods.
    The trajectories are plotted for the gene pairs with the highest correlation (ABCA3 and A1BG), and for the first two principal components.
    }
    \label{fig:main_result}
    \vspace{-1em}
\end{figure*}

\subsection{Unconditional Generation: Developmental Modeling of Embryonic Stem Cells}
\label{sec:exp_unconditional_generation}
\textbf{Data.}
Deciphering the developmental behavior of cells is the quintessential goal in the field of stem cell research \cite{alison2002introduction,zakrzewski2019stem}.
The experiment is conducted on scRNA-seq data of embryonic stem cells \cite{moon2019visualizing}, which is collected during the developmental stages over a period of 27 days, split into five phases: $t_0$ (days 0-3), $t_1$ (days 6-9), $t_2$ (days 12-15), $t_3$ (days 18-21), and $t_4$ (days 24-27).
Following the setting in \cite{tong2020trajectorynet,koshizuka2022neural}, gene expressions are (linearly) projected into a lower-dimensional space through principal component analysis (PCA)  \cite{wold1987principal} prior to conducting the experiments,
which also can be re-projected to the original space for evaluation.

\textbf{Evaluation and compared methods.}
To evaluate the (biological) validity of the proposed population guidance, we conduct model training using data from the terminal stages ($t_0, t_4$) without access to the intermediate ($t_1, t_2, t_3$), which are held out for evaluation.
Following the setting in \cite{tong2020trajectorynet,koshizuka2022neural}, the model is evaluated in the scenarios where the trajectories are generated based on $\hat{p}_{t_0}$ (referred as ``all-step" prediction), or based on  $\hat{p}_{t_{i-1}}$ for $\pi_{t_i}$ (``one-step"),
and the performance is quantified for the intermediate stages, based on the discrepancy in the  Wasserstein distance \cite{villani2009optimal,santambrogio2015optimal} between the predicted and the ground truth PCs.

The compared baselines include random expressions sampled from a non-informative uniform distribution, ODE-based approaches  OT-Flow \cite{onken2021ot} and  TrajectoryNet \cite{tong2020trajectorynet}, and
SDE-based approaches DMSB \cite{chen2023deep}, NeuralSDE \cite{li2020scalable,tzen2019neural} and NLSB \cite{koshizuka2022neural}.
The proposed CLSB falls under the category of SDE-based approaches.
We adopt neural SDEs for parametrizing dynamics, with the guidance weights tuned via grid search.

\textbf{Results (i). Population guidance leads to more accurate modeling of cell developmental dynamics.}
The results of developmental modeling of embryonic stem cells are shown in Tab. \ref{tab:uncondition_generation}.
Compared with the competitors, CLSB without individual guidance ($\alpha_\text{ind}=0$) attains the lowest average rank, which indicates it replicates developmental gene expressions in better alignment with the ground truth.
This improvement is particularly evident in the most challenging stage of $t_2$, which is far from both end points observed at $t_0$ and $t_4$.
This demonstrates the effectiveness of the proposed population guidance in the heterogeneous systems of cell clusters.
We also observe that in comparison to ODE-based methods, SDE-based approaches perform better, aligning with the inherently stochastic and diffuse nature of cell expression priors \cite{koshizuka2022neural}.

\textbf{(ii) Population guidance aids in generating cell states as a better proxy to the ground truth.}
The visualization of the simulated gene expressions and trajectories are shown in Fig. \ref{fig:main_result}.
Compared with individual-level guidance, population guidance not only improves the local accuracy of modeling for specific genes (rows 1-2, the distribution of two genes with the highest correlation), but also globally generates cell states that more effectively approximate the ground truth (rows 3-4, the distribution of the first two principal components).
The visualizations covering more gene pairs and principal components are depicted in Append. \ref{append:more_visualization}.

\textbf{(iii) Different population guidance strategies serve varied functions.}
We further carry out ablation studies to examine the contributions of three instantiated population objectives to overall performance, as detailed in Append. \ref{append:more_visualization} Tab. \ref{tab:ablation_studies}.
Interestingly, we find they serve varied functions.
Specifically, restraining the ``acceleration" of covariance ($k=2$, instantiation 2 \eqref{eq:covariance_acceleration}) provides more benefit in the early stage of development (i.e. $t_1$), and restraining the ``velocity" of covariance ($k=1$, instantiation 1 \eqref{eq:covariance_velocity}) provides more benefit in the later stages (i.e. $t_2, t_3$).
This observation could indicate that in nature, co-expression relations among genes undergo larger magnitude variations in early development stages, and tend to stabilize as development progresses.
The benefit of aligning with known gene-gene interactions ($k=0$, instantiation 3 \eqref{eq:covariance_potential}) is present across all stages, albeit modestly.

\begin{table*}[!t]
\begin{center}
\caption{Evaluation in the conditional generation scenario of dose-dependent cellular response prediction to chemical perturbations. Numbers ($\times$1e-3) indicate the mean and median Wasserstein distances for all genes, where lower values are preferable.
}
\vspace{-0.5em}
\label{tab:condition_generation}
\resizebox{0.95\textwidth}{!}{
\begin{tabular}{c | c c | c c | c c | c c }
    \toprule
    \multirow{2}{*}{Methods} & \multicolumn{2}{c|}{$t_1$} & \multicolumn{2}{c|}{$t_2$ (Most Challenging)} & \multicolumn{2}{c|}{$t_3$} & \multirow{2}{*}{\textbf{A.R.}} \\
    & Mean & Median & Mean & Median & Mean & Median & \\
    \midrule
    Random & 573.0$\pm$51.3 & 516.9$\pm$24.2 & 578.7$\pm$51.7 & 520.2$\pm$24.6 & 577.5$\pm$52.9 & 519.0$\pm$25.1 & 7.5 \\
    NeuralSDE(RandInit) & 529.9$\pm$73.4 & 578.7$\pm$46.2 & 536.5$\pm$73.4 & 592.5$\pm$45.3 & 536.3$\pm$75.8 & 585.7$\pm$53.0 & 7.5 \\
    VAE & 227.6$\pm$87.5 & 168.6$\pm$107.1 & 215.7$\pm$74.7 & 159.6$\pm$87.8 & 210.0$\pm$70.0 & 150.5$\pm$77.7 & 6.0 \\
    NeuralODE & 177.7$\pm$43.2 & 108.3$\pm$38.7 & 192.3$\pm$56.1 & 119.8$\pm$50.0 & 183.5$\pm$58.4 & 115.5$\pm$51.4 & 4.8 \\
    NeuralSDE & 170.1$\pm$40.8 & 102.0$\pm$44.8 & 183.0$\pm$53.2 & 117.3$\pm$57.1 & 182.3$\pm$63.4 & 117.8$\pm$55.0 & 4.1 \\
    NLSB(E) & \darkred{\textbf{78.6}}$\pm$35.1 & \darkred{\textbf{59.8}}$\pm$29.3 & 93.1$\pm$43.1 & \darkred{\textbf{70.0}}$\pm$35.8 & 104.0$\pm$47.9 & 75.9$\pm$39.6 & 2.3 \\
    \midrule
    CLSB($\alpha_\text{ind}>0$) & 79.3$\pm$36.4 & 61.0$\pm$29.6 & \darkred{\textbf{87.3}}$\pm$44.8 & \darkred{\textbf{67.0}}$\pm$34.6 & \darkred{\textbf{92.0}}$\pm$50.0 & \darkred{\textbf{69.4}}$\pm$38.1 & \darkred{\textbf{1.5}} \\
    CLSB($\alpha_\text{ind}=0$) & \darkred{\textbf{76.9}}$\pm$33.7 & \darkred{\textbf{60.9}}$\pm$27.4 & \darkred{\textbf{89.1}}$\pm$39.9 & 72.2$\pm$32.5 & \darkred{\textbf{93.3}}$\pm$43.2 & \darkred{\textbf{74.8}}$\pm$35.4 & \darkred{\textbf{2.1}} \\
    \bottomrule
\end{tabular}} \vspace{-1em}
\end{center}
\end{table*}

\subsection{Conditional Generation: Dose-Dependent Cellular Response Prediction to Chemical Perturbations}
\label{sec:exp_conditional_generation}

\textbf{Data.}
Examining cellular responses to chemical perturbations is one of the fundamental tasks in the drug discovery process \cite{dong2023causal,bunne2023learning}.
The experiment utilizes sciplex data from three cancer cell lines under different drug treatments \cite{srivatsan2020massively}, where data are collected for treatment doses of 10 nM, 100 nM, 1 $\mu$M, and 10 $\mu$M.
In this context, we consider the drug dose as pseudo-time (denoted as $t_1, t_2, t_3, t_4$, respectively), and define the gene expression in the absence of treatment as $t_0$.

\textbf{Evaluation and compared methods.}
We train our model using samples from the terminal stages ($t_0, t_4$), reserving samples from the intermediate stages ($t_1, t_2, t_3$) for evaluation. During inference, expressions are generated based on the state at $t_0$. Performance is then assessed more pragmatically based on the Wasserstein distance for each gene, which is further compared on the mean and median values.

The compared baselines include the uniform random expressions, VAE \cite{kingma2013auto}, NeuralODE \cite{onken2021ot}, NeuralSDE \cite{li2020scalable,tzen2019neural}, and NLSB \cite{koshizuka2022neural}.
For all methods, we additionally incorporate a molecular graph representation of the drug treatment using an expressive graph attention network \cite{velivckovic2017graph}, which is jointly trained with the generative model.

\textbf{Results (iv). Incorporation of population and individual-level guidance leads to more accurate prediction of perturbation effects.}
The results of dose-dependent cellular
response prediction to chemical perturbations are shown in Tab. \ref{tab:condition_generation}.
Compared with the competitors, CLSB incorporated with individual guidance ($\alpha_\text{ind}>0$) attains the lowest average rank, which indicates it replicates treated gene expressions in better alignment with the ground truth,
and the benefit of population guidance is presented in all the three stages.
This coincides the effectiveness of the proposed population guidance.
We also observe the similar phenomenon that SDE-based approaches outperform ODE-based approaches, and the classical VAE.

\section{Conclusions}
In this paper, we introduce a novel framework termed Correlational Lagrangian Schr\"odinger Bridge (CLSB), effectively addressing the challenges posed by restricted cross-sectional samples and the heterogeneous nature of individual particles. 
By shifting the focus of regularization from individual-level to population, CLSB acknowledges and leverages the inherent heterogeneity in systems to improve model generalizability.   
In developing CLSB, we address the technical challenges including (1) a new class of population regularizers capturing with the tractable formulation, (2) domain-informed instantiations, and (3) the integration of into data-driven generative models.
Numerically, we validate the superiority of CLSB in modeling cellular systems.


\section{Impact Statements}
This paper presents work whose goal is to advance the field of Machine Learning. There are many potential societal consequences of our work, none which we feel must be specifically highlighted here.





\bibliography{example_paper}
\bibliographystyle{icml2024}

\newpage
\appendix
\onecolumn

\section*{Appendix}

\section{Proof for Proposition 1}
\label{append:proof_for_proposition_1}
\textbf{Proposition 1.}
For revisit, correlational Lagrangian is defined as:
\begin{align*}
    L_\text{corr}(\pi_t, \widetilde{\mathcal{M}}, k) = \Big| \frac{\infm^k}{\infm t^k} \mathbb{E}_{\pi_t} [ \prod_{(j, m) \in \widetilde{\mathcal{M}}} (\mathrm{x}_{t, [j]})^m ] \Big|^2.
\end{align*}
For $k=1$, it admits the analytical expression in Eq. \eqref{eq:corr_lagr_1}, if for the set of functions $\mathcal{H} = \{ h(\vct{x}) \pi_t(\vct{x}) \vct{v}_t(\vct{x}),$
$\pi_t(\vct{x}) \vct{D}(\vct{x}) \nabla h(\vct{x}),$
$\pi_t(\vct{x}) \nabla^\top\vct{D}_t(\vct{x}) h(\vct{x}),$
$h(\vct{x}) \vct{D}_t(\vct{x}) \nabla \pi_t(\vct{x}) \}$ ($\theta$ is omitted for simplicity) that $h(\vct{x}) = \prod_{(j, m) \in \widetilde{\mathcal{M}}} (\mathrm{x}_{t, [j]})^m$, $\vct{D}_t(\vct{x}) = \vct{\Sigma}_t(\vct{x}) \vct{\Sigma}_t^\top(\vct{x})$, it satisfies:
\begin{itemize}[leftmargin=*] \vspace{-0.7em}
    \item Continuity. $h' \in \mathcal{H}$ is continuously differentiable w.r.t. $\vct{x}$;
    \vspace{-0.7em}
    \item Light tail. The probability density function $\pi_t(\vct{x})$ is characterized by tails that are sufficiently light, such that $\oint_{S_\infty} h'(\vct{x})  \cdot \infm \vct{a} = 0$ for $h' \in \mathcal{H}$, where $\vct{a}$ is the outward pointing unit normal on the boundary at infinity $S_\infty$.
\end{itemize} \vspace{-0.7em}
For $k \ge 2$, it admits the analytical expression in Eq. \eqref{eq:corr_lagr_k} in an iterative manner, similarly, if for the set of functions $\mathcal{H} = \{ h(\vct{x}) \pi_t(\vct{x}) \vct{v}_t(\vct{x}),$
$\pi_t(\vct{x}) \vct{D}(\vct{x}) \nabla h(\vct{x}),$
$\pi_t(\vct{x}) \nabla^\top\vct{D}_t(\vct{x}) h(\vct{x}),$
$h(\vct{x}) \vct{D}_t(\vct{x}) \nabla \pi_t(\vct{x}) \}$ that $h(\vct{x}) = \circ_{\Upsilon_{<i, j>} \in \widetilde{\mathcal{S}}} \Upsilon_{<i, j>}( \vct{x} )$, $\forall \widetilde{\mathcal{S}} \in \mathcal{S}_{<k-1>}$, it satisfies:
\begin{itemize}[leftmargin=*] \vspace{-0.7em}
    \item Continuity. $h' \in \mathcal{H}$ is continuously differentiable w.r.t. $\vct{x}$;
    \vspace{-0.7em}
    \item Light tail. The probability density function $\pi_t(\vct{x})$ is characterized by tails that are sufficiently light, such that $\oint_{S_\infty} h'(\vct{x})  \cdot \infm \vct{a} = 0$ for $h' \in \mathcal{H}$.
\end{itemize} \vspace{-0.7em}

\textit{Proof}.
\textbf{Expression for $k=1$.}
For simplicity, we omit $\theta$ in notations that $\vct{v}_t(\vct{x}; \theta), \vct{\Sigma}_t(\vct{x}; \theta)$ are referred as $\vct{v}_t(\vct{x}), \vct{\Sigma}_t(\vct{x})$, respectively.
Denoting $h: \mathbb{R}^d \rightarrow \mathbb{R}$ as the mapping satisfying the continuity and light tail conditions, for $k=1$, we have:
\begin{align*}
     & \frac{\infm}{\infm t} \mathbb{E}_{\pi_t} [ h(\vct{\mathrm{x}}_t) ] \\
     \overset{(a)}{=} \; & \frac{\infm}{\infm t} \int h(\vct{x}) \pi_t(\vct{x}) \infm \vct{x} = \int h(\vct{x}) (\frac{\infm}{\infm t} \pi_t(\vct{x})) \infm \vct{x} \\
     \overset{(b)}{=} \; & \int h(\vct{x}) \Big( - \nabla \cdot ( \pi_t(\vct{x}) \vct{v}_t(\vct{x}) ) + \frac{1}{2} \nabla \cdot \nabla \cdot ( \pi_t(\vct{x}) \vct{\Sigma}_t(\vct{x}) \vct{\Sigma}_t^\top(\vct{x}) ) \Big) \infm \vct{x} \\
     = \; & \int \Big( - h(\vct{x}) \Big) \Big( \nabla \cdot ( \pi_t(\vct{x}) \vct{v}_t(\vct{x}) ) \Big) \infm \vct{x} + \int \Big( \frac{1}{2} h(\vct{x}) \Big) \Big( \nabla \cdot \nabla \cdot ( \pi_t(\vct{x}) \vct{\Sigma}_t(\vct{x}) \vct{\Sigma}_t^\top(\vct{x}) ) \Big) \infm \vct{x} \\
     \overset{(c)}{=} \; & \overbrace{\int \nabla \Big( h(\vct{x}) \Big) \cdot \Big( \pi_t(\vct{x}) \vct{v}_t(\vct{x}) \Big) \infm \vct{x}}^{\text{Part (i)}} + \overbrace{\int - \nabla \cdot \Big( h(\vct{x}) \pi_t(\vct{x}) \vct{v}_t(\vct{x}) \Big) \infm \vct{x}}^{\text{Part (ii)}} \\
     & + \overbrace{\int - \nabla \Big( \frac{1}{2} h(\vct{x}) \Big) \cdot \Big( \nabla \cdot ( \pi_t(\vct{x}) \vct{\Sigma}_t(\vct{x}) \vct{\Sigma}_t^\top(\vct{x}) ) \Big) \infm \vct{x}}^{\text{Part (iii)}} + \overbrace{\int \nabla \cdot \Big( \Big( \frac{1}{2} h(\vct{x}) \Big) \Big( \nabla \cdot ( \pi_t(\vct{x}) \vct{\Sigma}_t(\vct{x}) \vct{\Sigma}_t^\top(\vct{x}) ) \Big) \Big) \infm \vct{x}}^{\text{Part (iv)}},
\end{align*}
where (a) is attained through the standard definition of integration,
(b) results from the application of the Fokker-Planck equation \eqref{eq:fokker_planck} to substitute the time derivative term with the divergence term, and
(c) is achieved by applying integration by parts on the right-hand side (RHS) of Eq. (b).
Next, we solve the four parts on RHS of Eq. (c).
For part (i), we have:
\begin{align*}
    & \int \nabla \Big( h(\vct{x}) \Big) \cdot \Big( \pi_t(\vct{x}) \vct{v}_t(\vct{x}) \Big) \infm \vct{x} \\
    = \; & \int \Big( \nabla h(\vct{x}) \cdot \vct{v}_t(\vct{x}) \Big) \pi_t(\vct{x}) \infm \vct{x} \\
    = \; & \mathbb{E}_{\pi_t} [ \nabla h(\vct{x}) \cdot \vct{v}_t(\vct{x}) ].
\end{align*}
For part (ii), we have:
\begin{align*}
    & \int - \nabla \cdot \Big( h(\vct{x}) \pi_t(\vct{x}) \vct{v}_t(\vct{x}) \Big) \infm \vct{x} \\
    \overset{(a)}{=} \; & - \oint_{S_\infty} \Big( h(\vct{x}) \pi_t(\vct{x}) \vct{v}_t(\vct{x}) \Big) \cdot \infm \vct{a} \\
    \overset{(b)}{=} \; & 0,
\end{align*}
where (a) is accomplished through the application of Gauss's theorem \cite{temple1936gauss}, that $\vct{a}$ is the outward pointing unit normal at each point on the boundary at infinity $S_\infty$, under the satisfaction of the continuity condition, and (b) is achieved by considering the light tail condition.

For part (iii), we have:
\begin{align*}
    & \int \nabla \Big( h(\vct{x}) \Big) \cdot \Big( \nabla \cdot ( \pi_t(\vct{x}) \vct{\Sigma}_t(\vct{x}) \vct{\Sigma}_t^\top(\vct{x}) ) \Big) \infm \vct{x} \\
    \overset{(a)}{=} \; & \int \nabla \cdot \Big( \pi_t(\vct{x}) \vct{\Sigma}_t(\vct{x}) \vct{\Sigma}_t^\top(\vct{x}) \nabla  h(\vct{x}) \Big) \infm \vct{x} - \int \Big(\nabla \nabla^\top h(\vct{x})\Big) \bigcdot \Big( \pi_t(\vct{x}) \vct{\Sigma}_t(\vct{x}) \vct{\Sigma}_t^\top(\vct{x}) \Big) \infm \vct{x} \\
    \overset{(b)}{=} \; & \oint_{S_\infty} \Big( \pi_t(\vct{x}) \vct{\Sigma}_t(\vct{x}) \vct{\Sigma}_t^\top(\vct{x}) \nabla  h(\vct{x}) \Big) \cdot \infm \vct{a} - \mathbb{E}_{\pi_t}[\Big(\nabla \nabla^\top h(\vct{x})\Big) \bigcdot \Big( \vct{\Sigma}_t(\vct{x}) \vct{\Sigma}_t^\top(\vct{x}) \Big)] \\
    \overset{(c)}{=} \; & - \mathbb{E}_{\pi_t}[\Big(\nabla \nabla^\top h(\vct{x})\Big) \bigcdot \Big( \vct{\Sigma}_t(\vct{x}) \vct{\Sigma}_t^\top(\vct{x}) \Big)],
\end{align*}
where (a) is achieved by applying integration by parts,
(b) is accomplished through the application of Gauss's theorem under the satisfaction of the continuity condition, and
(c) is achieved by considering the light tail condition.

For part (iv), we have:
\begin{align*}
    & \int \nabla \cdot \Big( \Big( h(\vct{x}) \Big) \Big( \nabla \cdot ( \pi_t(\vct{x}) \vct{\Sigma}_t(\vct{x}) \vct{\Sigma}_t^\top(\vct{x}) ) \Big) \Big) \infm \vct{x} \\
    \overset{(a)}{=} \; & \int \nabla \cdot \Big( \Big( h(\vct{x}) \Big) \Big( \nabla \cdot ( \vct{\Sigma}_t(\vct{x}) \vct{\Sigma}_t^\top(\vct{x}) ) \Big) \pi_t(\vct{x}) \Big) \infm \vct{x} + \int \nabla \cdot \Big( \Big( h(\vct{x}) \Big) \Big( \nabla \pi_t(\vct{x}) \cdot ( \vct{\Sigma}_t(\vct{x}) \vct{\Sigma}_t^\top(\vct{x}) ) \Big) \Big) \infm \vct{x} \\
    \overset{(b)}{=} \; & \oint_{S_\infty} \Big( \Big( h(\vct{x}) \Big) \Big( \nabla \cdot ( \vct{\Sigma}_t(\vct{x}) \vct{\Sigma}_t^\top(\vct{x}) ) \Big) \pi_t(\vct{x}) \Big) \cdot \infm \vct{a} + \oint_{S_\infty} \Big( \Big( h(\vct{x}) \Big) \Big( \nabla \pi_t(\vct{x}) \cdot ( \vct{\Sigma}_t(\vct{x}) \vct{\Sigma}_t^\top(\vct{x}) ) \Big) \Big) \cdot \infm \vct{a} \\
    \overset{(c)}{=} \; & 0,
\end{align*}
where (a) is achieved by applying the product rule,
(b) is accomplished through the application of Gauss's theorem under the satisfaction of the continuity condition, and
(c) is achieved by considering the light tail condition.

By combining them and setting $h(\vct{x}) = \prod_{(j, m) \in \widetilde{\mathcal{M}}} (\mathrm{x}_{[j]})^m$, we eventually have:
\begin{align*}
    L_\text{corr}(\pi_t, \widetilde{\mathcal{M}}, 1) = \Big| \mathbb{E}_{\pi_t} [ \nabla \Big( \prod_{(j, m) \in \widetilde{\mathcal{M}}} (\mathrm{x}_{t, [j]})^m \Big) \cdot \vct{v}_t(\vct{\mathrm{x}}_t) ] + \frac{1}{2} \mathbb{E}_{\pi_t} [ (\nabla \nabla^\top \Big( \prod_{(j, m) \in \widetilde{\mathcal{M}}} (\mathrm{x}_{t, [j]})^m \Big)) \bigcdot (\vct{\Sigma}_t(\vct{\mathrm{x}}_t) \vct{\Sigma}^\top_t(\vct{\mathrm{x}}_t)) ] \Big|^2.
\end{align*}

\textbf{Expression for $k \ge 2$.}
We present a more general form of correlational Lagrangian, by
denoting the ordered sequence of operators $\widetilde{\mathcal{S}} = \{..., \Upsilon_{<i, j>}, ...\}$ that $\Upsilon \in \{\nabla, \nabla \nabla^\top\}, i \in \mathbb{Z}_>, j \in \mathbb{Z}_>$ such that
\begin{align*}
    \Upsilon_{<i, j>} (\vct{x}) = \frac{\infm^i}{\infm t^i} \Upsilon( h(\vct{x}) ) \# \frac{\infm^j}{\infm t^j} \gamma(\vct{x}), \quad\quad \# = \Big\{ \begin{smallmatrix} \cdot, \text{ if } \Upsilon = \nabla \\
    \bigcdot, \text{ else if } \Upsilon = \nabla \nabla^\top \end{smallmatrix},
    \gamma(\vct{x}) = \Big\{ \begin{smallmatrix} \vct{v}_t(\vct{x}), \text{ if } \Upsilon = \nabla \\
    \vct{\Sigma}_t(\vct{x}) \vct{\Sigma}^\top_t(\vct{x}), \text{ else if } \Upsilon = \nabla \nabla^\top \end{smallmatrix},
\end{align*}

and denote the function $\Gamma(\cdot)$ operating on the ordered sequence $\widetilde{\mathcal{S}}$ such that
\begin{align*}
\Gamma(\widetilde{\mathcal{S}}) = c_{\widetilde{\mathcal{S}}} \mathbb{E}_{\pi_t} [ \circ_{\Upsilon_{<i, j>} \in \widetilde{\mathcal{S}}} \Upsilon_{<i, j>}( \vct{\mathrm{x}}_t ) ], \quad\quad c_{\widetilde{\mathcal{S}}} = 2^{-|\{\Upsilon_{<i, j>}: \Upsilon_{<i, j>} \in \widetilde{\mathcal{S}}, \Upsilon = \nabla \nabla^\top\}|},
\end{align*}
and then we can rewrite correlational Lagrangian for the $k=1$ case:
\begin{align*}
    L_\text{corr}(\pi_t, \widetilde{\mathcal{M}}, 1) = \Big| \Gamma(\{\nabla_{<0, 0>}\}) + \Gamma(\{(\nabla \nabla^\top)_{<0, 0>}\} ) \Big|^2.
\end{align*}
We further denote $\mathcal{S}_{<k>}$ as the set of $\widetilde{\mathcal{S}}$ used to compute $L_\text{corr}(\pi_t, \widetilde{\mathcal{M}}, k)$, e.g., $\mathcal{S}_{<1>} = \{ \{\nabla_{<0, 0>}\}, \{(\nabla \nabla^\top)_{<0, 0>}\} \}$ according to the above formulation.
Thus, for $k \ge 2$, we have:
\begin{align*}
    L_\text{corr}(\pi_t, \widetilde{\mathcal{M}}, k) = \Big| \sum_{\widetilde{\mathcal{S}} \in \mathcal{S}_{<k-1>}} \frac{\infm}{\infm t} \Gamma( \widetilde{\mathcal{S}}  )  \Big|^2.
\end{align*}
To calculate this general formulation, denoting $\widetilde{\mathcal{S}} = \widetilde{\mathcal{S}}' \cup \{\Upsilon'_{<i', j'>}\}$, we utilize the following equation:
\begin{align*}
    & \frac{\infm}{\infm t} \Gamma( \widetilde{\mathcal{S}} ) = \frac{\infm}{\infm t} \Gamma( \widetilde{\mathcal{S}}' \cup \{\Upsilon'_{<i', j'>}\} ) \\
    \overset{(a)}{=} \; & c_{\widetilde{\mathcal{S}}} \frac{\infm}{\infm t} \mathbb{E}_{\pi_t} [ \circ_{\Upsilon_{<i, j>} \in \widetilde{\mathcal{S}}} \Upsilon_{<i, j>}( \vct{\mathrm{x}}_t ) ] \\
    \overset{(b)}{=} \; & c_{\widetilde{\mathcal{S}}} \frac{\infm}{\infm t} \int \frac{\infm^{i'}}{\infm t^{i'}} \Upsilon'\Big( \circ_{\Upsilon_{<i, j>} \in \widetilde{\mathcal{S}}'} \Upsilon_{<i, j>}( \vct{x} ) \Big) \#' \Big( \frac{\infm^{j'}}{\infm t^{j'}} \gamma'(\vct{x}) \Big) \pi_t(\vct{x}) \infm \vct{x} \\
    \overset{(c)}{=} \; & c_{\widetilde{\mathcal{S}}} \int \frac{\infm^{i'+1}}{\infm t^{i'+1}} \Upsilon'\Big( \circ_{\Upsilon_{<i, j>} \in \widetilde{\mathcal{S}}'} \Upsilon_{<i, j>}( \vct{x} ) \Big) \#' \Big( \frac{\infm^{j'}}{\infm t^{j'}} \gamma'(\vct{x}) \Big) \pi_t(\vct{x}) \infm \vct{x} \\
    & + c_{\widetilde{\mathcal{S}}} \int \frac{\infm^{i'}}{\infm t^{i'}} \Upsilon'\Big( \circ_{\Upsilon_{<i, j>} \in \widetilde{\mathcal{S}}'} \Upsilon_{<i, j>}( \vct{x} ) \Big) \#' \Big( \frac{\infm^{j'+1}}{\infm t^{j'+1}} \gamma'(\vct{x}) \Big) \pi_t(\vct{x}) \infm \vct{x} \\
    & + c_{\widetilde{\mathcal{S}}} \int \frac{\infm^{i'}}{\infm t^{i'}} \Upsilon'\Big( \circ_{\Upsilon_{<i, j>} \in \widetilde{\mathcal{S}}'} \Upsilon_{<i, j>}( \vct{x} ) \Big) \#' \Big( \frac{\infm^{j'}}{\infm t^{j'}} \gamma'(\vct{x}) \Big) \Big( \frac{\infm}{\infm t} \pi_t(\vct{x}) \Big) \infm \vct{x} \\
    \overset{(d)}{=} \; & \Gamma( \widetilde{\mathcal{S}}' \cup \{\Upsilon'_{<i'+1, j'>}\} ) + \Gamma( \widetilde{\mathcal{S}}' \cup \{\Upsilon'_{<i', j'+1>}\} ) + c_{\widetilde{\mathcal{S}}} \int \Big( \circ_{\Upsilon_{<i, j>} \in \widetilde{\mathcal{S}}} \Upsilon_{<i, j>}( \vct{x} ) \Big) \Big( \frac{\infm}{\infm t} \pi_t(\vct{x}) \Big) \infm \vct{x},
\end{align*}
where (a, b) is established through definitions, (c) is realized by applying the product rule, and (d) is also derived from standard definitions.
Denoting $h(\vct{x}) = \circ_{\Upsilon_{<i, j>} \in \widetilde{\mathcal{S}}} \Upsilon_{<i, j>}( \vct{x} )$, we have:
\begin{align*}
    & c_{\widetilde{\mathcal{S}}} \int h(\vct{x}) ( \frac{\infm}{\infm t} \pi_t(\vct{x}) ) \infm \vct{x} \\
    \overset{(a)}{=} \; & c_{\widetilde{\mathcal{S}}} \mathbb{E}_{\pi_t} [ \nabla h(\vct{x}) \cdot \vct{v}_t(\vct{\mathrm{x}}_t) ] + \frac{c_{\widetilde{\mathcal{S}}}}{2} \mathbb{E}_{\pi_t} [ (\nabla \nabla^\top ( h(\vct{x}) )) \bigcdot (\vct{\Sigma}_t(\vct{\mathrm{x}}_t) \vct{\Sigma}^\top_t(\vct{\mathrm{x}}_t)) ] \\
    \overset{(b)}{=} \; & \Gamma( \widetilde{\mathcal{S}} \cup \{\nabla_{<0, 0>}\} ) + \Gamma( \widetilde{\mathcal{S}} \cup \{(\nabla \nabla^\top)_{<0, 0>}\} ),
\end{align*}
where (a) follows the same derivation of correlational Lagrangian for $k=1$, under the satisfaction of the continuity condition and light tail conditions,
and (b) is established through definitions.

By combining them, we eventually have:
\begin{align*}
    L_\text{corr}(\pi_t, \widetilde{\mathcal{M}}, k) = \Big| \sum_{\substack{\widetilde{\mathcal{S}} \in \mathcal{S}_{<k-1>}, \\ \widetilde{\mathcal{S}} = \widetilde{\mathcal{S}}' \cup \{\Upsilon'_{<i', j'>}\} }} \Gamma(\widetilde{\mathcal{S}}' \cup \Upsilon'_{<i'+1, j'>}) + \Gamma(\widetilde{\mathcal{S}}' \cup \Upsilon'_{<i', j'+1>}) + \Gamma(\widetilde{\mathcal{S}}  \cup \nabla_{<0, 0>}) + \Gamma(\widetilde{\mathcal{S}}  \cup (\nabla \nabla^\top)_{<0, 0>}) \Big|^2.
\end{align*}

\section{More Related Works}
\label{append:related_works}

\textbf{Modeling population dynamics with machine learning.}
A significant body of research has been dedicated to modeling population dynamics using data-driven approaches. This includes the development of continuous normalizing flows \cite{chen2018neural,mathieu2020riemannian}, which model the dynamics through ordinary differential equations (ODEs). Furthermore, an advancement of neural ODEs, namely neural SDEs, has been introduced to capture both drift and diffusion processes using neural networks \cite{li2020scalable,tzen2019neural}. In scenarios where ground truth trajectories are inaccessible, regularization strategies for flows have been developed. These strategies emphasize enforcing constraints on the motion of individual trajectories. Examples include the regularization of kinetic energy and its Jacobian \cite{tong2020trajectorynet,finlay2020train}, as well as the inclusion of dual terms derived from the Hamilton–Jacobi–Bellman equation \cite{koshizuka2022neural,onken2021ot}, aiming to guide the model towards realistic dynamic behaviors.

\textbf{Developmental modeling of embryonic stem cells.}
The modeling of embryonic stem cell development represents a cutting-edge intersection of developmental biology, computational science, and systems biology \cite{alison2002introduction,zakrzewski2019stem}. This field aims to unravel the complex processes governing the differentiation and proliferation of embryonic stem cells into the diverse cell types that form an organism. Given the foundational role of these processes in understanding both normal development and various diseases, developmental modeling of embryonic stem cells has garnered significant interest.
At its core, developmental modeling seeks to simulate and predict the dynamic behavior of stem cells as they progress through various stages of development. This involves mapping the intricate pathways that lead to cell fate decisions, a challenge that requires sophisticated computational models and deep biological insights.

\textbf{Dose-dependent cellular response prediction to chemical perturbations.}
The prediction of dose-dependent cellular responses to chemical perturbations is pivotal in pharmacology, toxicology, and systems biology \cite{dong2023causal,bunne2023learning}. It aims to understand how cells react to varying concentrations of chemical compounds, which is crucial for drug development, safety assessment, and personalized medicine. This field combines quantitative biology, computational modeling, and high-throughput experimental techniques to map out the intricate cellular mechanisms activated or inhibited by drugs and other chemical agents at different doses.
At the heart of dose-dependent cellular response prediction is the need to accurately model the complex, nonlinear interactions between chemical perturbations and cellular pathways. This involves determining the specific dose at which a chemical agent begins to have a biological effect (the threshold), the range over which the response changes (the dynamic range), and the dose causing maximal response (the ceiling).

\section{Stability of Genetic Co-Expression Relations}
\label{append:coexpression_stability}
\begin{figure}[!htb] 
    \centering 
    \includegraphics[width=0.9\linewidth]{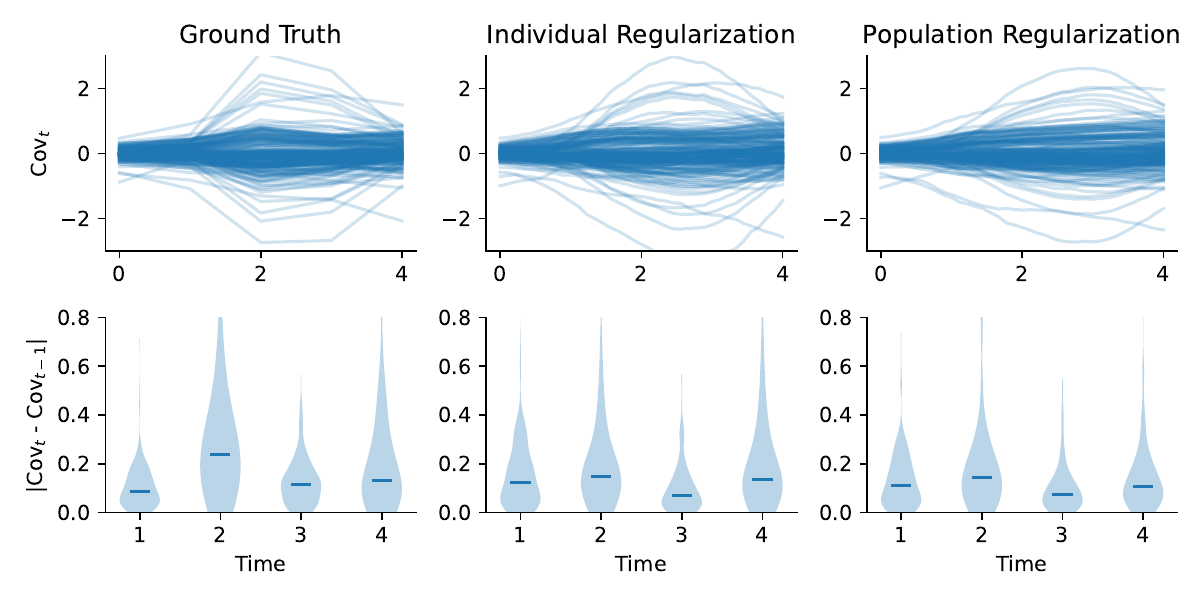}
    \caption{Visualization of temporal variations of the covariance of embryonic stem cell expression.
    The first row of figures presents direct plots of the covariance at time $t$, while the second row displays violin plots illustrating the differences between time $t$ and $t-1$.}
    \label{fig:covariance}
\end{figure}

\textbf{Stability of genetic co-expression relations.}
The majority of co-expression relationships among gene pairs remain relatively stable over time, as evidenced by the first column of Fig. \ref{fig:covariance}. Population guidance effectively preserves this stability, a feature that is often lost with individual-level guidance by comparing between the second and third columns of Fig. \ref{fig:covariance}.

\section{Formulation to Restrain ``Acceleration" of Standardized Covariance}
\label{append:acceleration_of_standardized_covariance}
The formulation to restrain the ``acceleration" of standardized covariance is expressed as:
\begin{align*}
    & \sum_{\widetilde{\mathcal{M}} \in \mathcal{M}_\text{cov}}  L_\text{std-corr}(\pi_t, \widetilde{\mathcal{M}}, 2) = \Big\lVert \frac{\infm^2}{\infm t^2} \Big( \mathbb{E}_{\pi_t}[ \vct{\mathrm{x}}_t \vct{\mathrm{x}}_t^\top ] - \mathbb{E}_{\pi_t}[ \vct{\mathrm{x}}_t] \mathbb{E}_{\pi_t}^\top[ \vct{\mathrm{x}}_t ] \Big) \Big\rVert^2_\mathsf{F} \\
    = \; & \Big\lVert \overbrace{\mathbb{E}_{\pi_t}\Big[\vct{\mathrm{x}}_t (\frac{\infm}{\infm t}\vct{v}_t(\vct{\mathrm{x}}_t))^\top + (\frac{\infm}{\infm t}\vct{v}_t(\vct{\mathrm{x}}_t)) \vct{\mathrm{x}}_t^\top + \frac{1}{2} \frac{\infm}{\infm t}( \vct{\Sigma}_t(\vct{\mathrm{x}}_t) \vct{\Sigma}^\top_t(\vct{\mathrm{x}}_t) ) \Big]}^{\text{Collection of } \Gamma(\widetilde{\mathcal{S}}' \cup \Upsilon'_{<i', j'+1>}) \text{ terms in Eq. \eqref{eq:corr_lagr_k}}} \notag \\
    & - \mathbb{E}_{\pi_t}[\vct{\mathrm{x}}_t] \mathbb{E}_{\pi_t}^\top[\frac{\infm}{\infm t}\vct{v}_t(\vct{\mathrm{x}}_t)] - \mathbb{E}_{\pi_t}[\frac{\infm}{\infm t}\vct{v}_t(\vct{\mathrm{x}}_t)] \mathbb{E}_{\pi_t}^\top[\vct{\mathrm{x}}_t] \quad\quad\quad\quad\quad\quad\quad\quad + \overbrace{\vct{0}}^{\Gamma(\widetilde{\mathcal{S}}' \cup \Upsilon'_{<i'+1, j'>})} \\
    & + \overbrace{\mathbb{E}_{\pi_t}\Big[ \vct{\mathrm{x}}_t (\nabla(\vct{v}_t(\vct{\mathrm{x}}_t))\vct{v}_t(\vct{\mathrm{x}}_t))^\top + (\nabla(\vct{v}_t(\vct{\mathrm{x}}_t))\vct{v}_t(\vct{\mathrm{x}}_t)) \vct{\mathrm{x}}_t^\top + 2 \vct{v}_t(\vct{\mathrm{x}}_t) \vct{v}_t(\vct{\mathrm{x}}_t)^\top + \frac{1}{2} \nabla(\vct{\Sigma}_t(\vct{\mathrm{x}}_t) \vct{\Sigma}^\top_t(\vct{\mathrm{x}}_t))_{\underline{i_1 i_2 i_3}} \vct{v}_t^{\underline{i_3}}(\vct{\mathrm{x}}_t) \Big]}^{\Gamma(\widetilde{\mathcal{S}} \cup \nabla_{<0, 0>})} \\
    & - \mathbb{E}_{\pi_t}[\vct{\mathrm{x}}_t] \mathbb{E}_{\pi_t}^\top[\nabla(\vct{v}_t(\vct{\mathrm{x}}_t))\vct{v}_t(\vct{\mathrm{x}}_t)] - \mathbb{E}_{\pi_t}[\nabla(\vct{v}_t(\vct{\mathrm{x}}_t))\vct{v}_t(\vct{\mathrm{x}}_t)] \mathbb{E}_{\pi_t}^\top[\vct{\mathrm{x}}_t] - 2 \mathbb{E}_{\pi_t}[\vct{v}_t(\vct{\mathrm{x}}_t)] \mathbb{E}_{\pi_t}^\top[\vct{v}_t(\vct{\mathrm{x}}_t)] \\
    & + \overbrace{\mathbb{E}_{\pi_t}\Big[ \nabla(\vct{v}_t(\vct{\mathrm{x}}_t)) \vct{\Sigma}_t(\vct{\mathrm{x}}_t) \vct{\Sigma}^\top_t(\vct{\mathrm{x}}_t) + \vct{\Sigma}_t(\vct{\mathrm{x}}_t) \vct{\Sigma}^\top_t(\vct{\mathrm{x}}_t) \nabla^\top(\vct{v}_t(\vct{\mathrm{x}}_t)) + \frac{1}{4} \nabla \nabla^\top(\vct{\Sigma}_t(\vct{\mathrm{x}}_t) \vct{\Sigma}^\top_t(\vct{\mathrm{x}}_t))_{\underline{i_1 i_2 i_3 i_4}} (\vct{\Sigma}_t(\vct{\mathrm{x}}_t) \vct{\Sigma}^\top_t(\vct{\mathrm{x}}_t))^{\underline{i_3 i_4}} }^{\Gamma(\widetilde{\mathcal{S}} \cup \nabla \nabla^\top_{<0, 0>})} \\
    & + \frac{1}{2} \vct{\mathrm{x}}_t ( \nabla \nabla^\top(\vct{v}_t(\vct{\mathrm{x}}_t))_{\underline{i_1 i_2 i_3}} (\vct{\Sigma}_t(\vct{\mathrm{x}}_t) \vct{\Sigma}^\top_t(\vct{\mathrm{x}}_t))^{\underline{i_2 i_3}} )^\top + \frac{1}{2} (\nabla \nabla^\top(\vct{v}_t(\vct{\mathrm{x}}_t))_{\underline{i_1 i_2 i_3}} (\vct{\Sigma}_t(\vct{\mathrm{x}}_t) \vct{\Sigma}^\top_t(\vct{\mathrm{x}}_t))^{\underline{i_2 i_3}}) \vct{\mathrm{x}}_t^\top \Big] \\
    & - \frac{1}{2} \mathbb{E}_{\pi_t}[\vct{\mathrm{x}}_t] \mathbb{E}_{\pi_t}^\top[\nabla \nabla^\top(\vct{v}_t(\vct{\mathrm{x}}_t))_{\underline{i_1 i_2 i_3}} (\vct{\Sigma}_t(\vct{\mathrm{x}}_t) \vct{\Sigma}^\top_t(\vct{\mathrm{x}}_t))^{\underline{i_2 i_3}}] - \frac{1}{2} \mathbb{E}_{\pi_t}[\nabla \nabla^\top(\vct{v}_t(\vct{\mathrm{x}}_t))_{\underline{i_1 i_2 i_3}} (\vct{\Sigma}_t(\vct{\mathrm{x}}_t) \vct{\Sigma}^\top_t(\vct{\mathrm{x}}_t))^{\underline{i_2 i_3}}] \mathbb{E}_{\pi_t}^\top[\vct{\mathrm{x}}_t] \Big\rVert^2_\mathsf{F},
\end{align*}
where $\vct{C} = \vct{A}_{\underline{ij}} \vct{B}^{\underline{jk}}$ adopts the Einstein notation for tensor operations that $\vct{C}_{[i,k]} = \sum_j \vct{A}_{[i,j]} \vct{B}_{[j,k]}$.

\section{Form of Covariance Potential}
\label{append:form_of_potential_function}
Denoting $\vct{D} = \mathbb{E}_{\pi_t}[ \vct{\mathrm{x}}_t \vct{\mathrm{x}}_t^\top ] - \mathbb{E}_{\pi_t}[ \vct{\mathrm{x}}_t] \mathbb{E}_{\pi_t}^\top[ \vct{\mathrm{x}}_t ]$ $\widetilde{\vct{D}} \in \mathbb{R}^{d \times d}, \widetilde{\vct{D}}_{[i,j]} = \Big\{ \begin{smallmatrix} 0, \text{ if } i \neq j \\
    1 / \sqrt{\vct{D}_{[i,j]}}, \text{ else,} \end{smallmatrix}$,
the designated form of potential covariance is expressed as:
\begin{align*}
    \sum_{\widetilde{\mathcal{M}} \in \mathcal{M}_\text{cov}} L_\text{corr}(\pi_t, \widetilde{\mathcal{M}}, 0) = \Big\lVert \widetilde{\vct{D}}^\top \vct{D} \widetilde{\vct{D}} - \vct{Y} \Big\rVert^2_\mathsf{F}.
\end{align*}

\section{More Visualizations and Results}
\label{append:more_visualization}

\begin{figure}[!htb] 
    \centering 
    \includegraphics[width=1\linewidth]{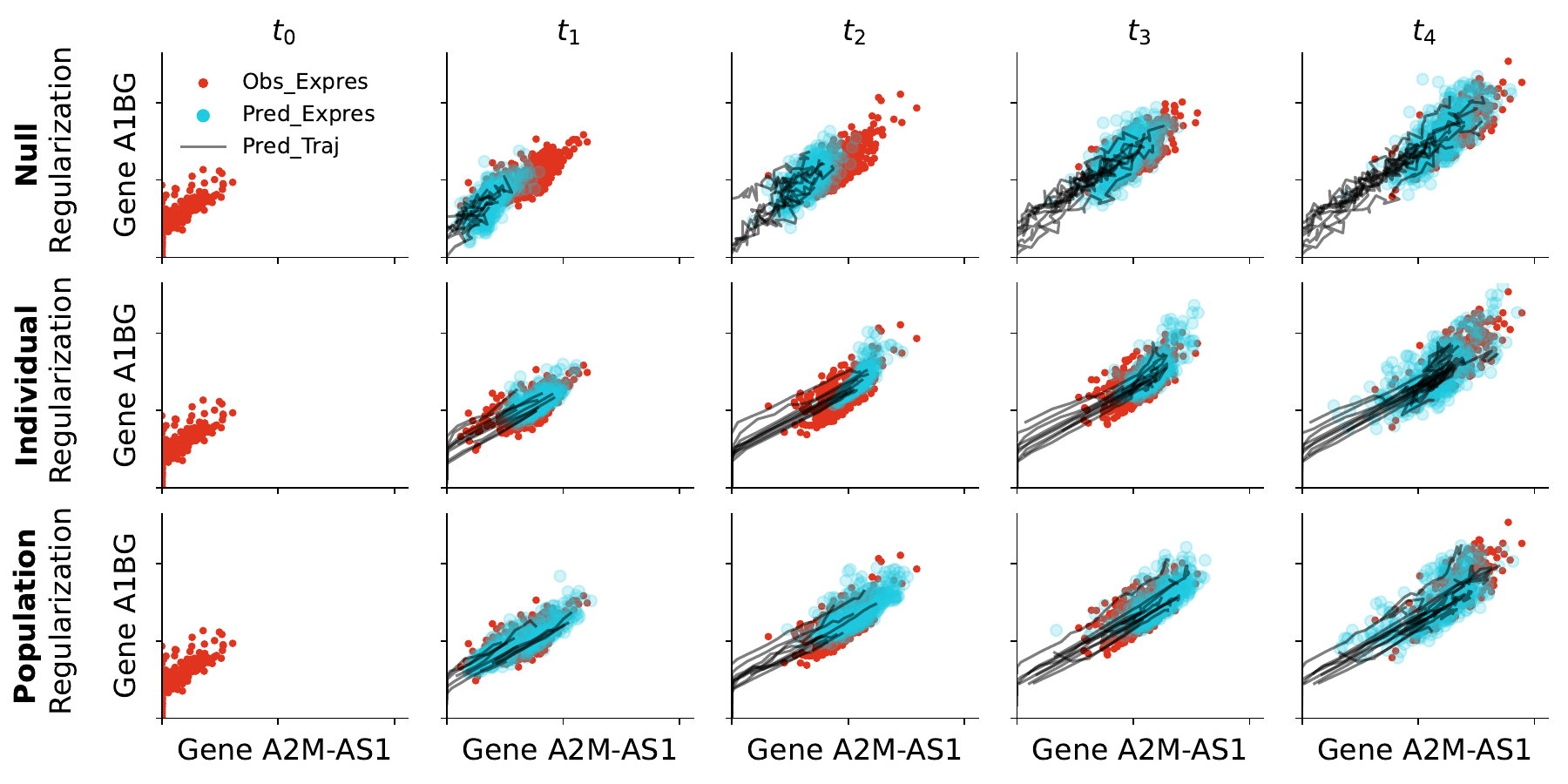}
    \caption{Visualization of local dynamics for genes A2M-AS1 and A1BG.}
\end{figure}

\begin{figure}[!htb] 
    \centering 
    \includegraphics[width=1\linewidth]{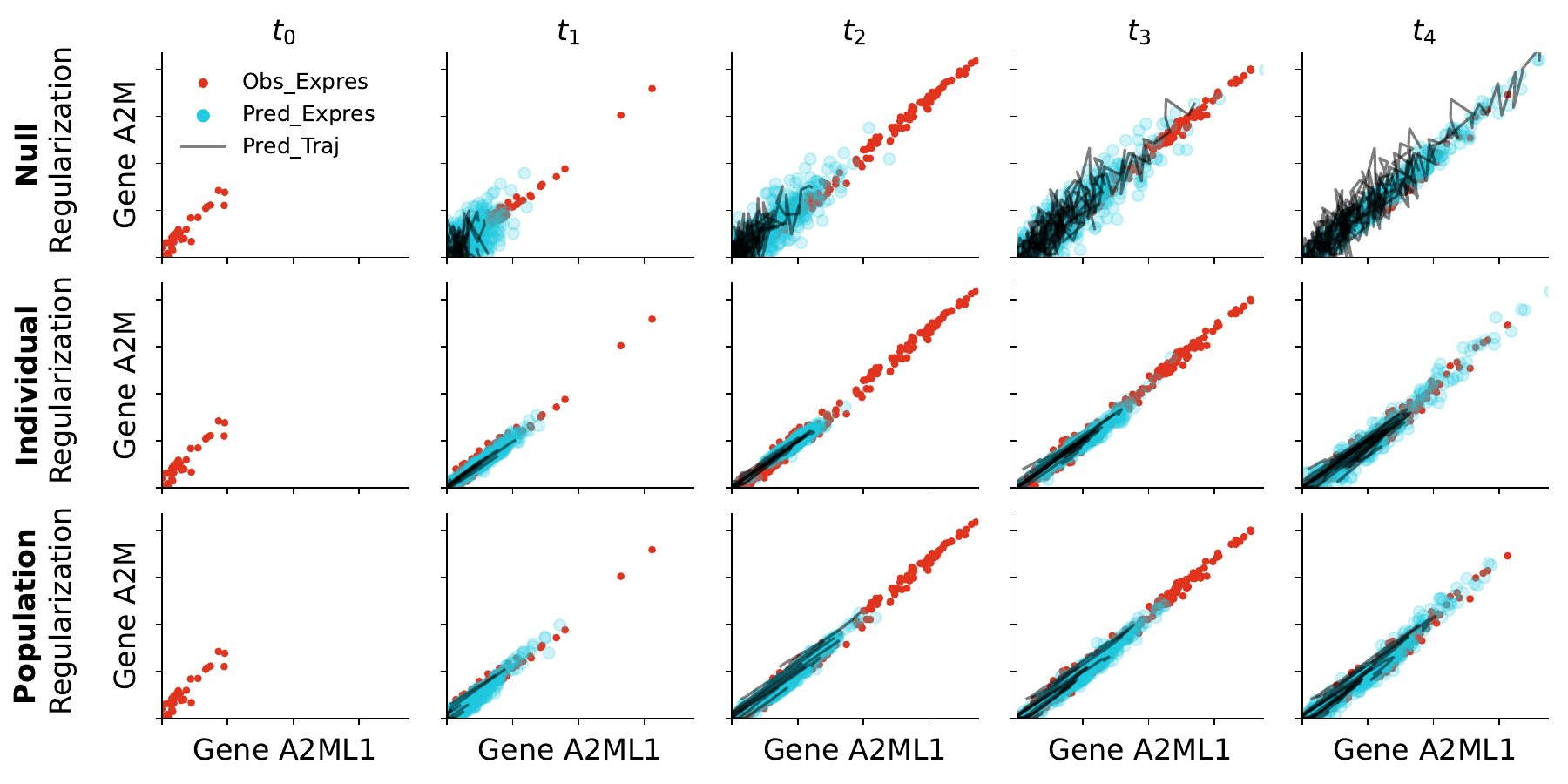}
    \caption{Visualization of local dynamics for genes A2ML1 and A2M.}
\end{figure}

\begin{figure}[!htb] 
    \centering 
    \includegraphics[width=1\linewidth]{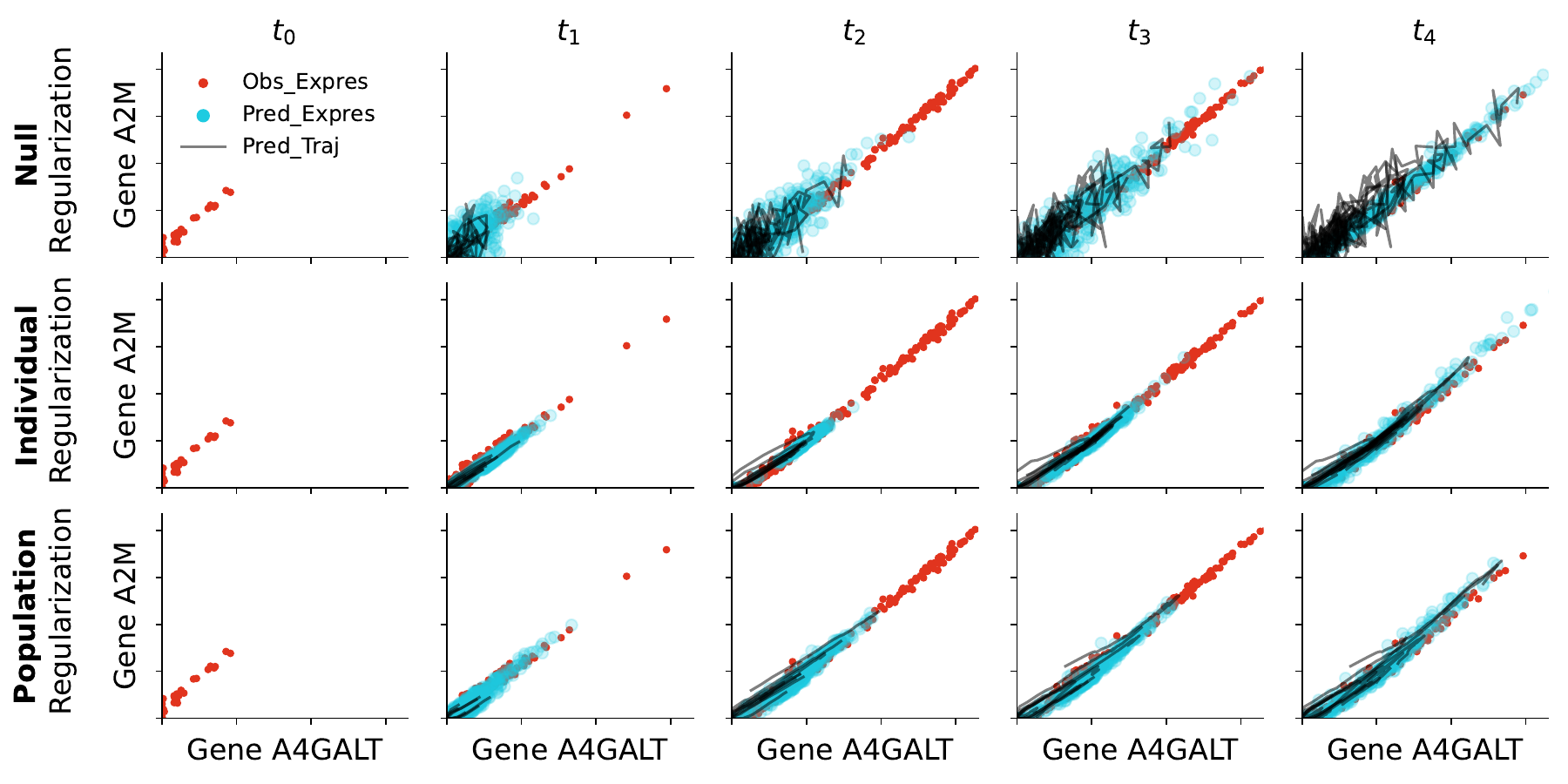}
    \caption{Visualization of local dynamics for genes A4GALT and A2M.}
\end{figure}

\begin{figure}[!htb] 
    \centering 
    \includegraphics[width=1\linewidth]{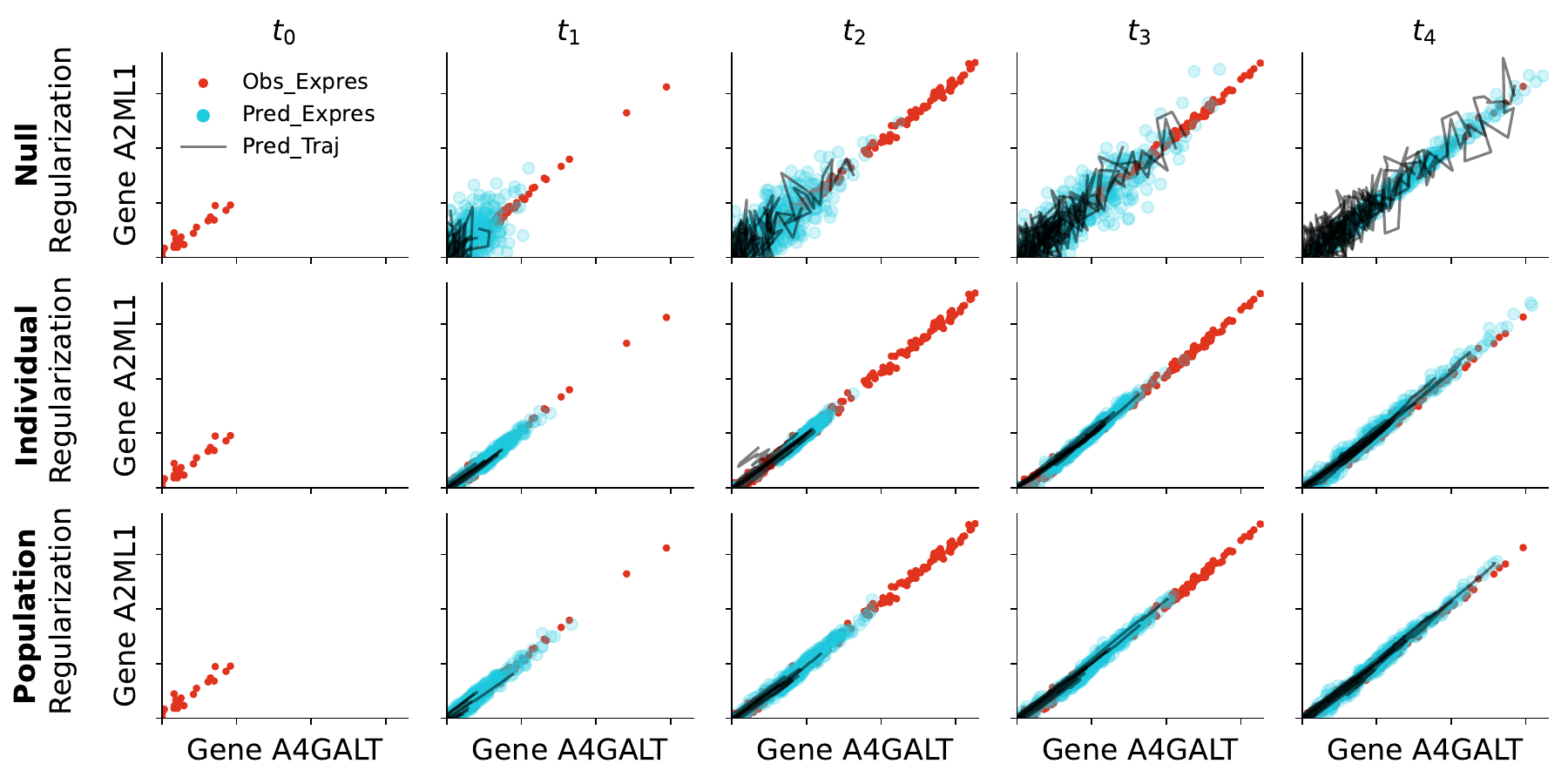}
    \caption{Visualization of local dynamics for genes A4GALT and A2ML1.}
\end{figure}

\begin{figure}[!htb] 
    \centering 
    \includegraphics[width=1\linewidth]{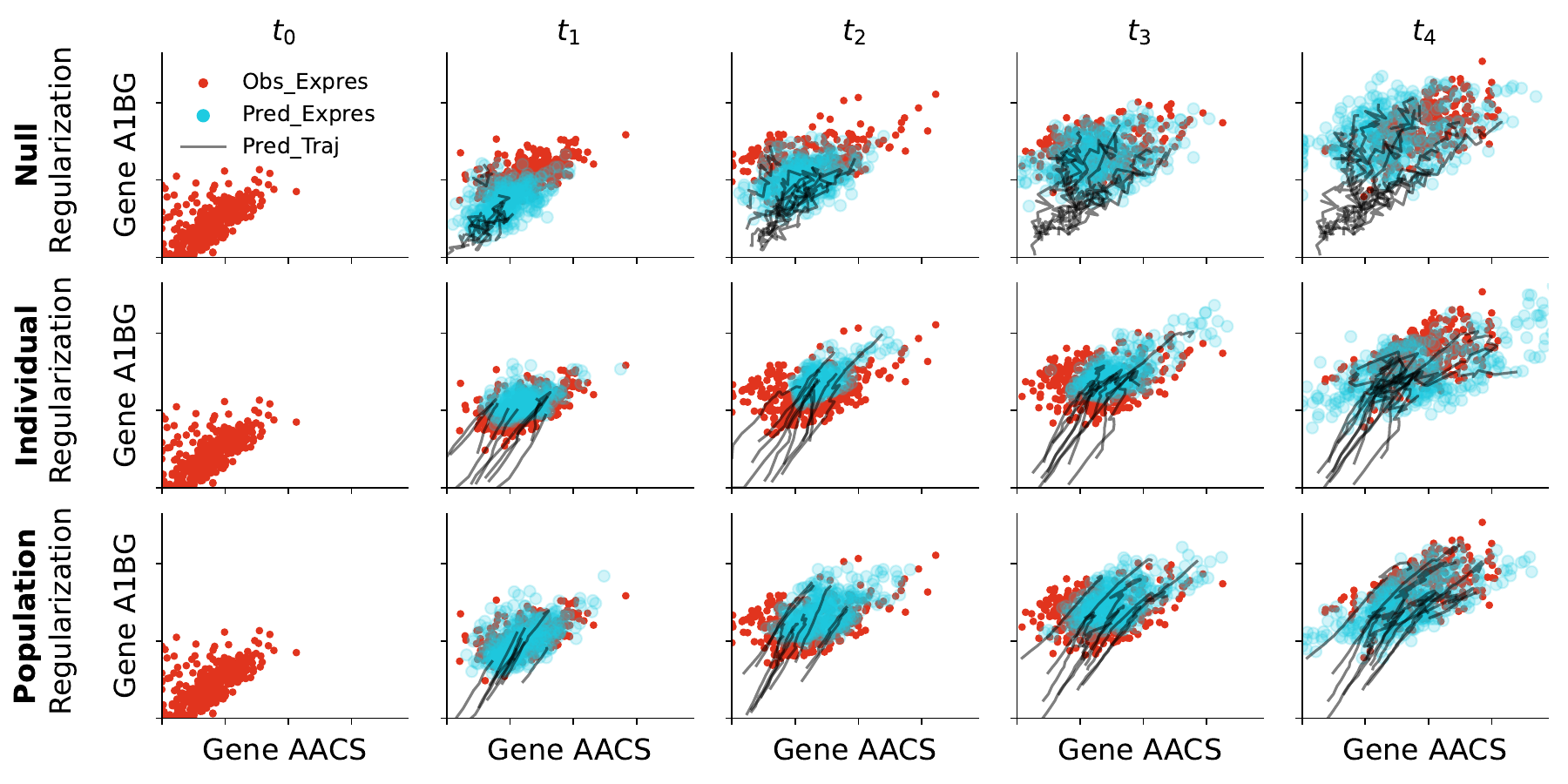}
    \caption{Visualization of local dynamics for genes AACS and A1BG.}
\end{figure}

\begin{figure}[!htb] 
    \centering 
    \includegraphics[width=1\linewidth]{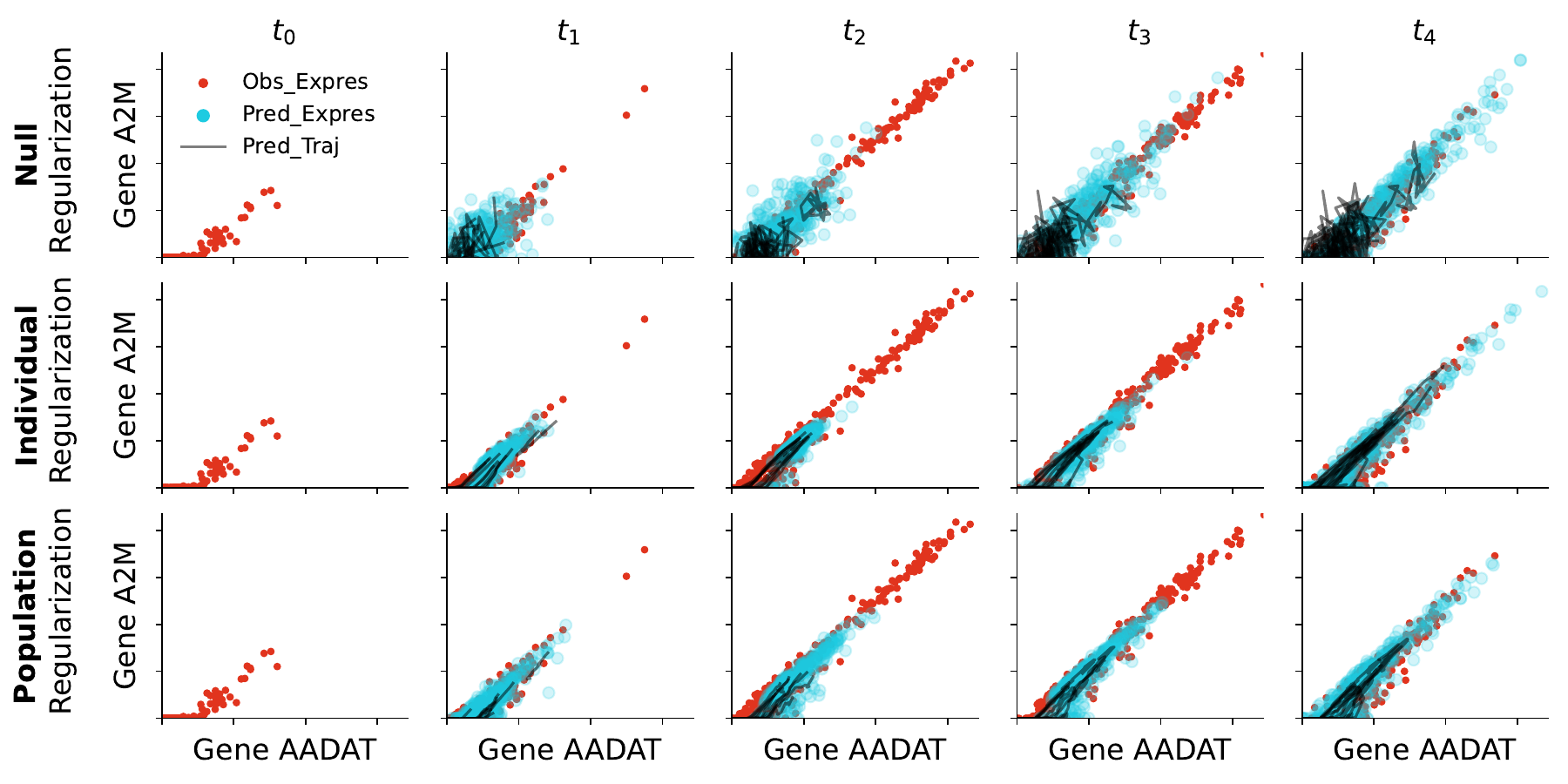}
    \caption{Visualization of local dynamics for genes AADAT and A2M.}
\end{figure}

\begin{figure}[!htb] 
    \centering 
    \includegraphics[width=1\linewidth]{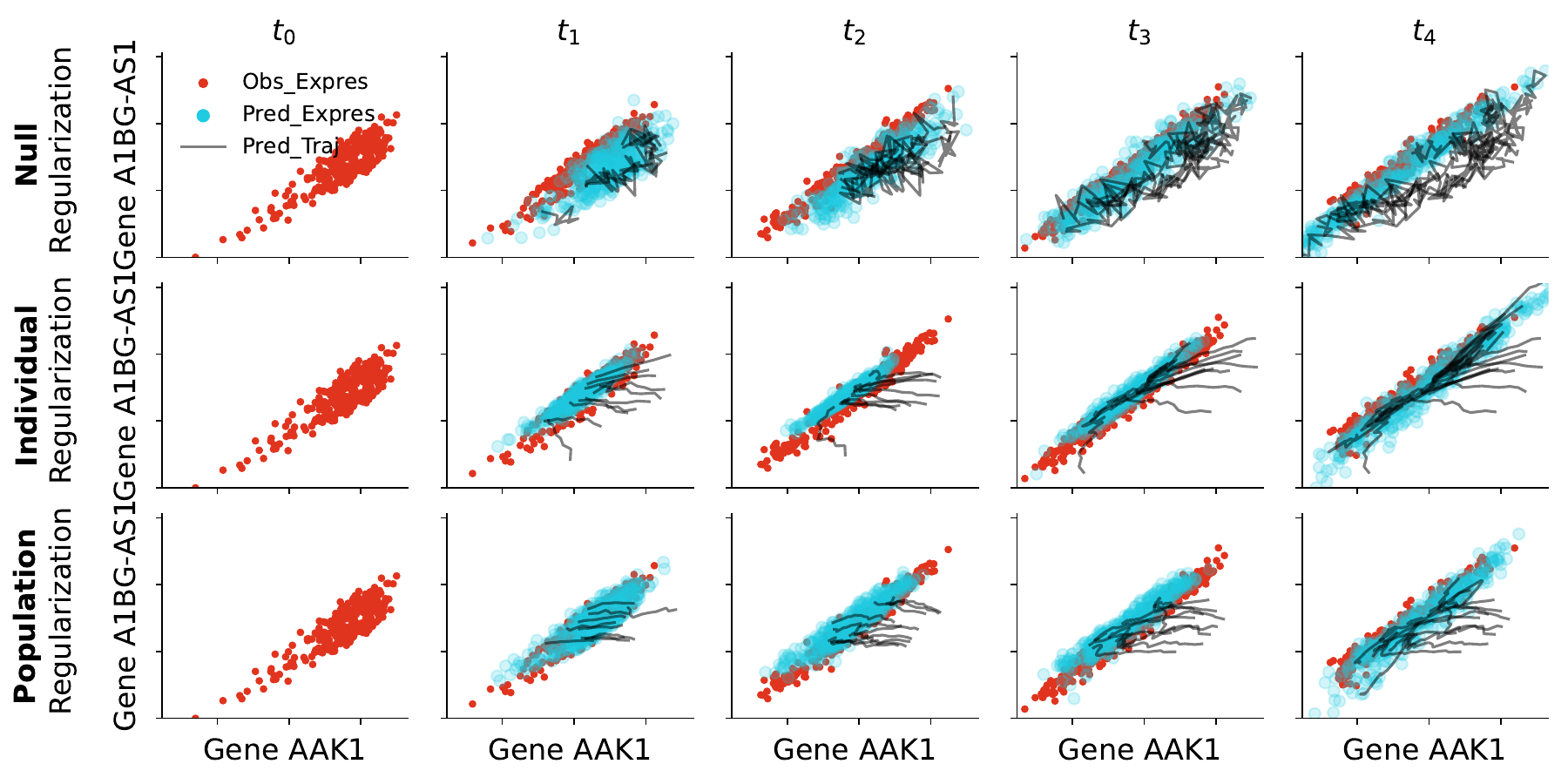}
    \caption{Visualization of local dynamics for genes AAK1 and A1BG-AS1.}
\end{figure}

\begin{figure}[!htb] 
    \centering 
    \includegraphics[width=1\linewidth]{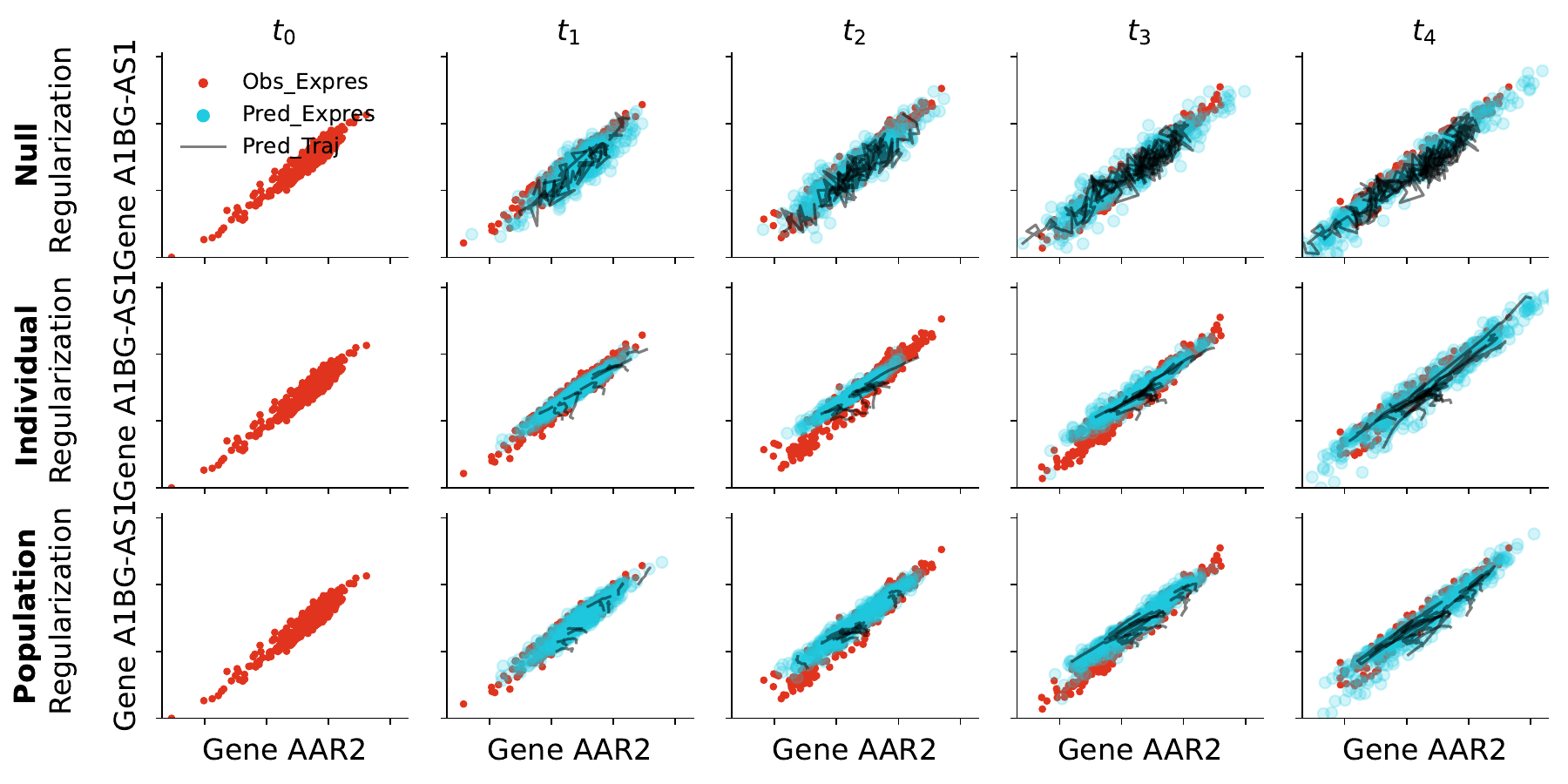}
    \caption{Visualization of local dynamics for genes AAR2 and A1BG-AS1.}
\end{figure}

\begin{figure}[!htb] 
    \centering 
    \includegraphics[width=1\linewidth]{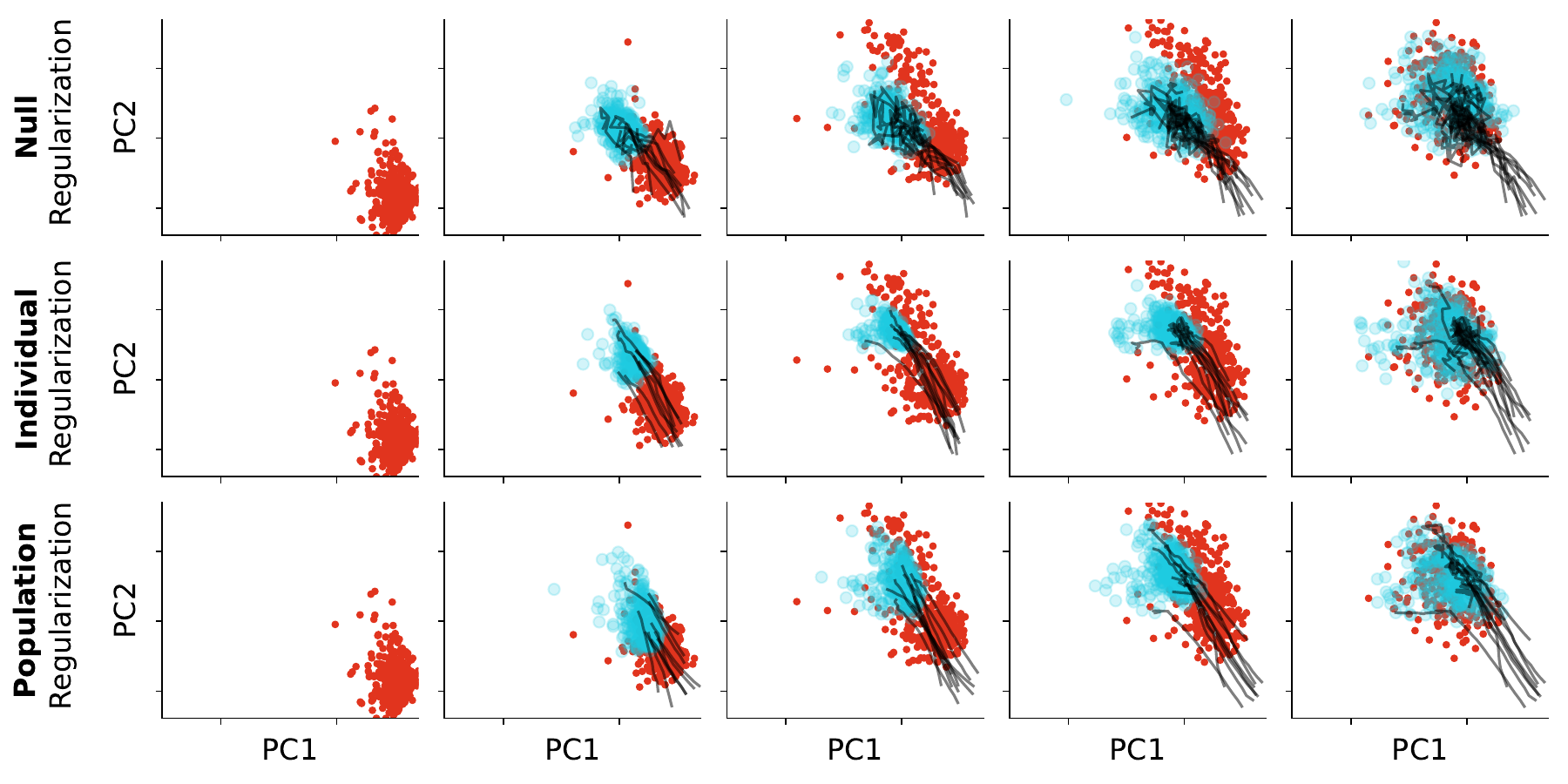}
    \caption{Visualization of global dynamics for principle components 1 and 2.}
\end{figure}

\begin{figure}[!htb] 
    \centering 
    \includegraphics[width=1\linewidth]{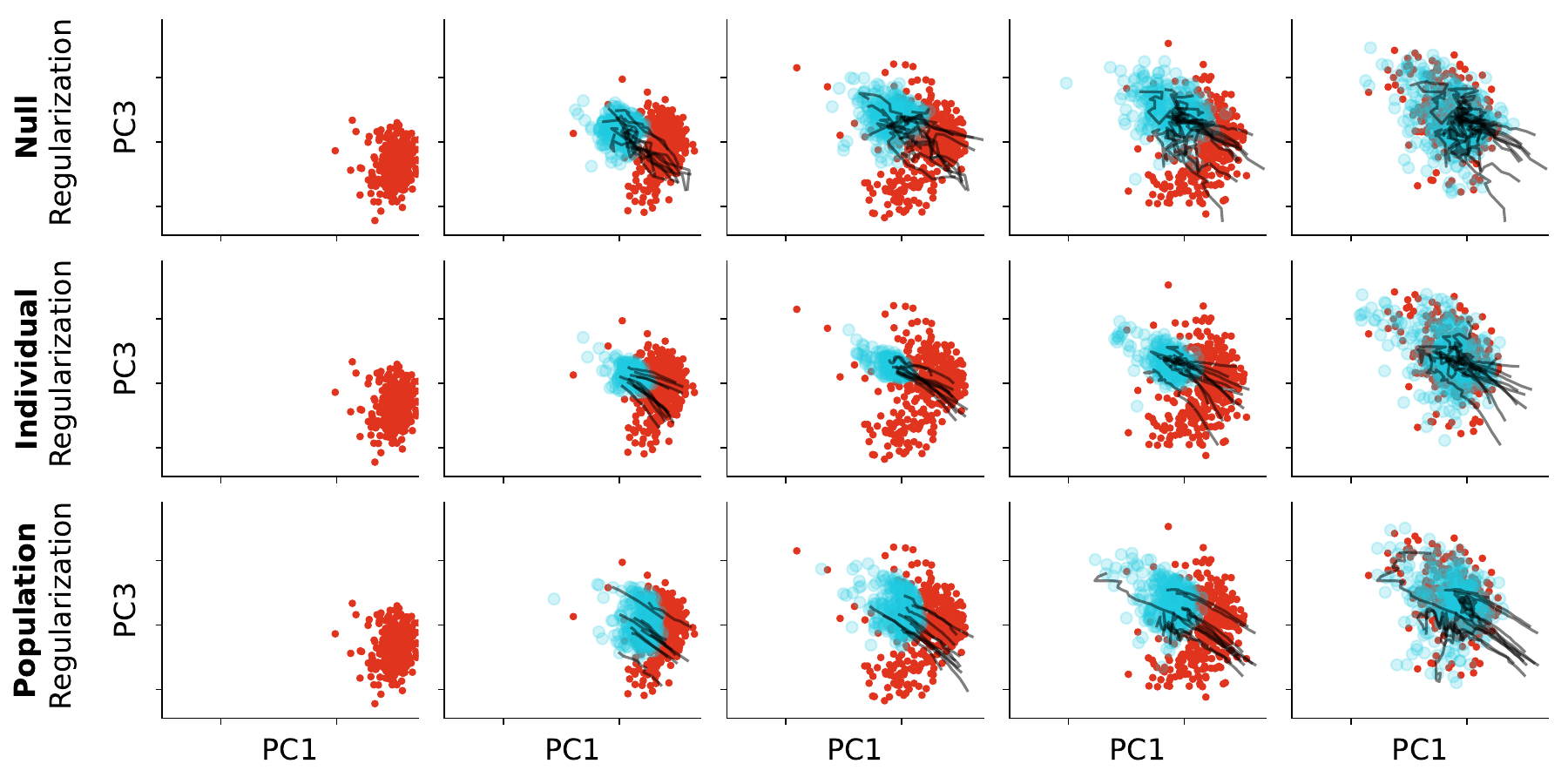}
    \caption{Visualization of global dynamics for principle components 1 and 3.}
\end{figure}

\begin{figure}[!htb] 
    \centering 
    \includegraphics[width=1\linewidth]{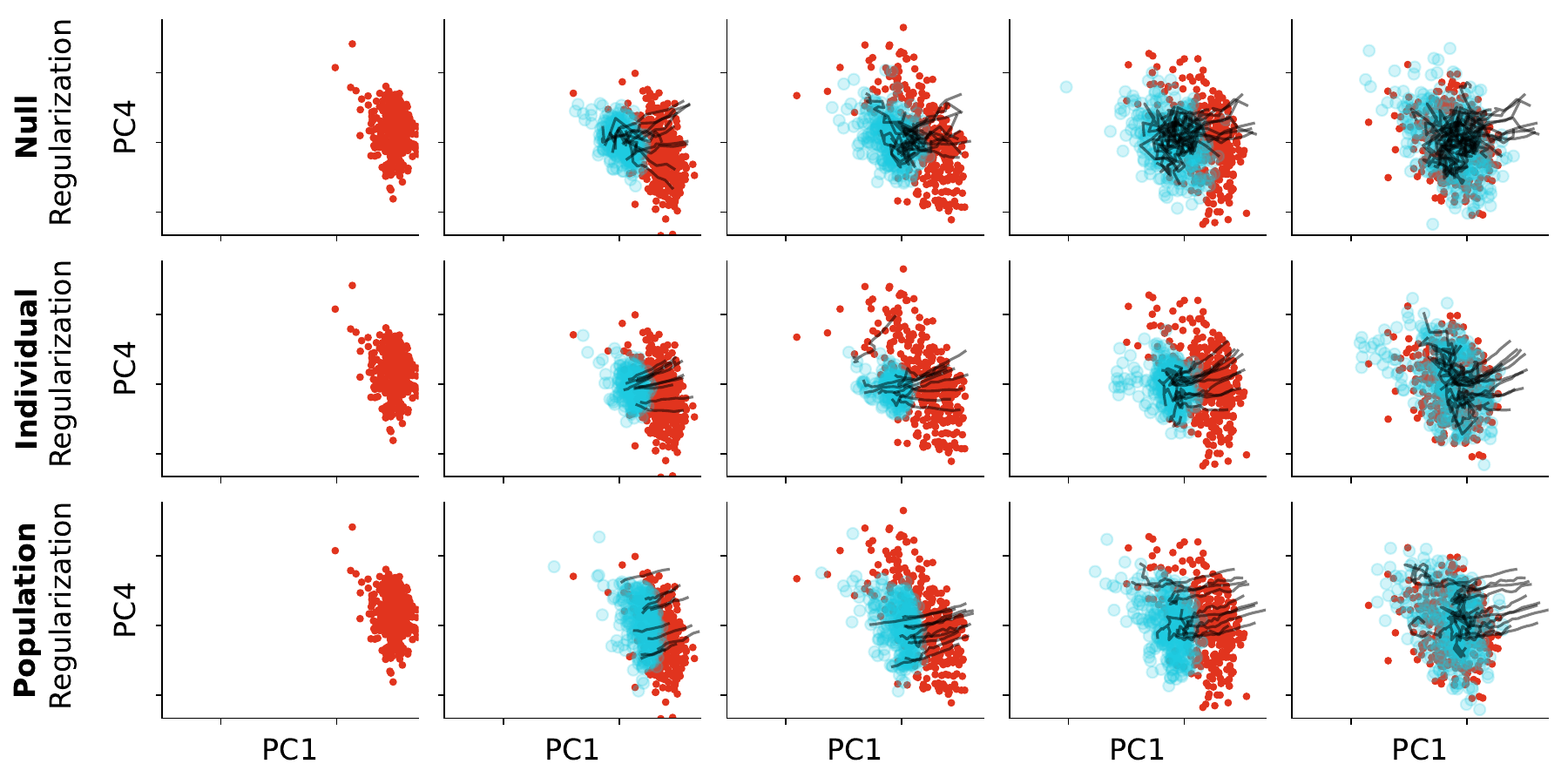}
    \caption{Visualization of global dynamics for principle components 1 and 4.}
\end{figure}

\begin{figure}[!htb] 
    \centering 
    \includegraphics[width=1\linewidth]{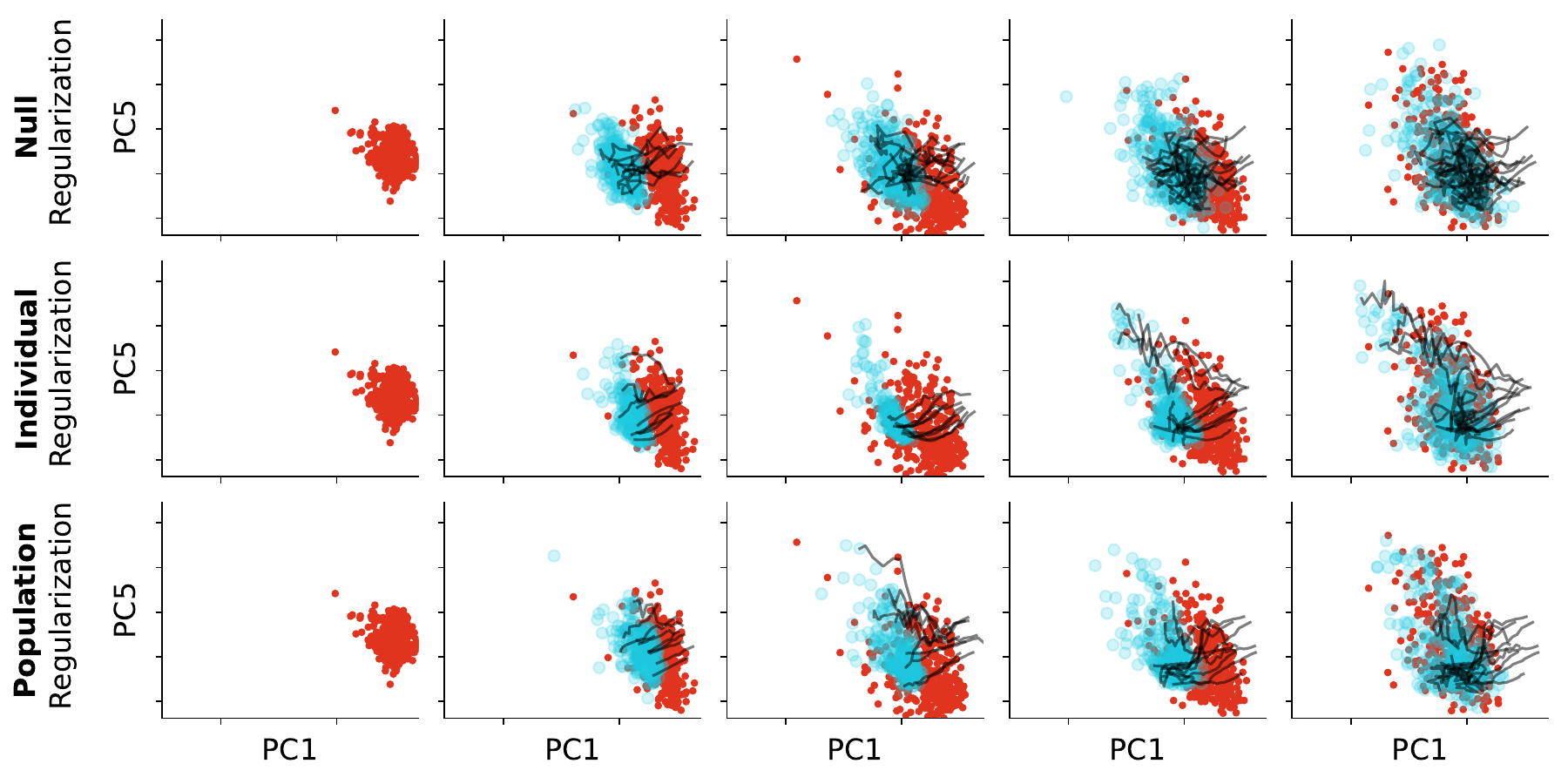}
    \caption{Visualization of global dynamics for principle components 1 and 5.}
\end{figure}

\begin{figure}[!htb] 
    \centering 
    \includegraphics[width=1\linewidth]{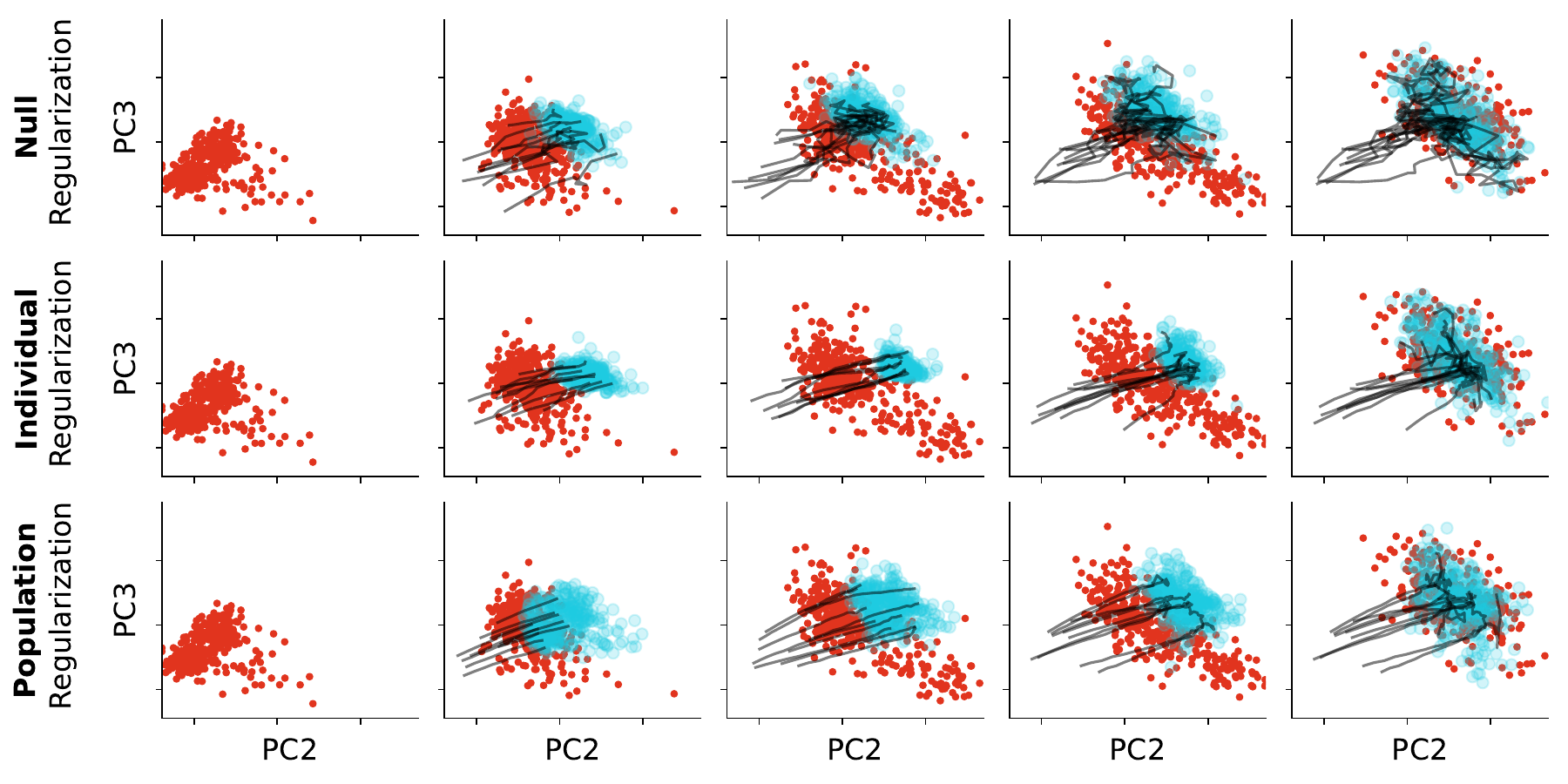}
    \caption{Visualization of global dynamics for principle components 2 and 3.}
\end{figure}

\begin{figure}[!htb] 
    \centering 
    \includegraphics[width=1\linewidth]{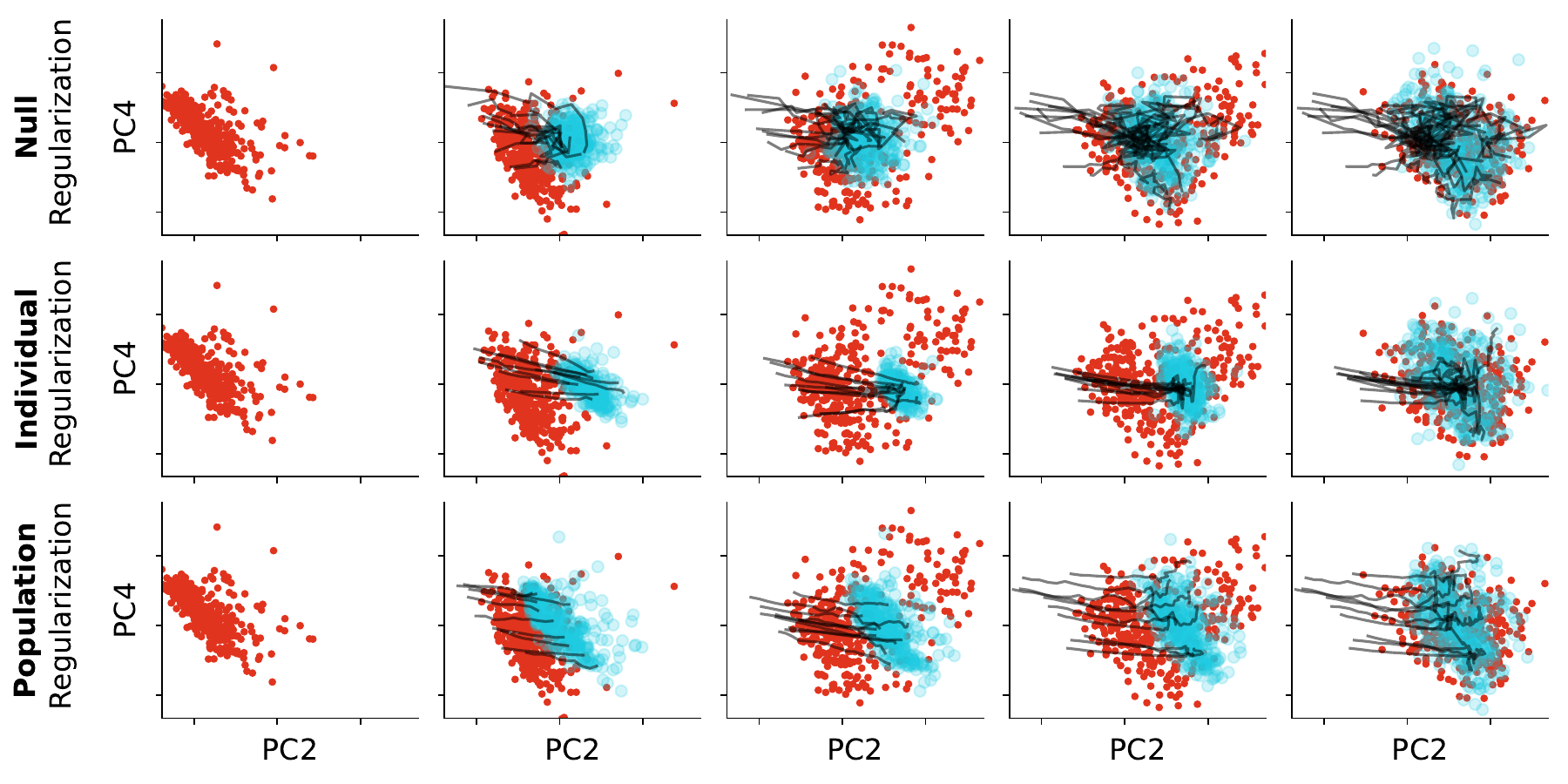}
    \caption{Visualization of global dynamics for principle components 2 and 4.}
\end{figure}

\begin{figure}[!htb] 
    \centering 
    \includegraphics[width=1\linewidth]{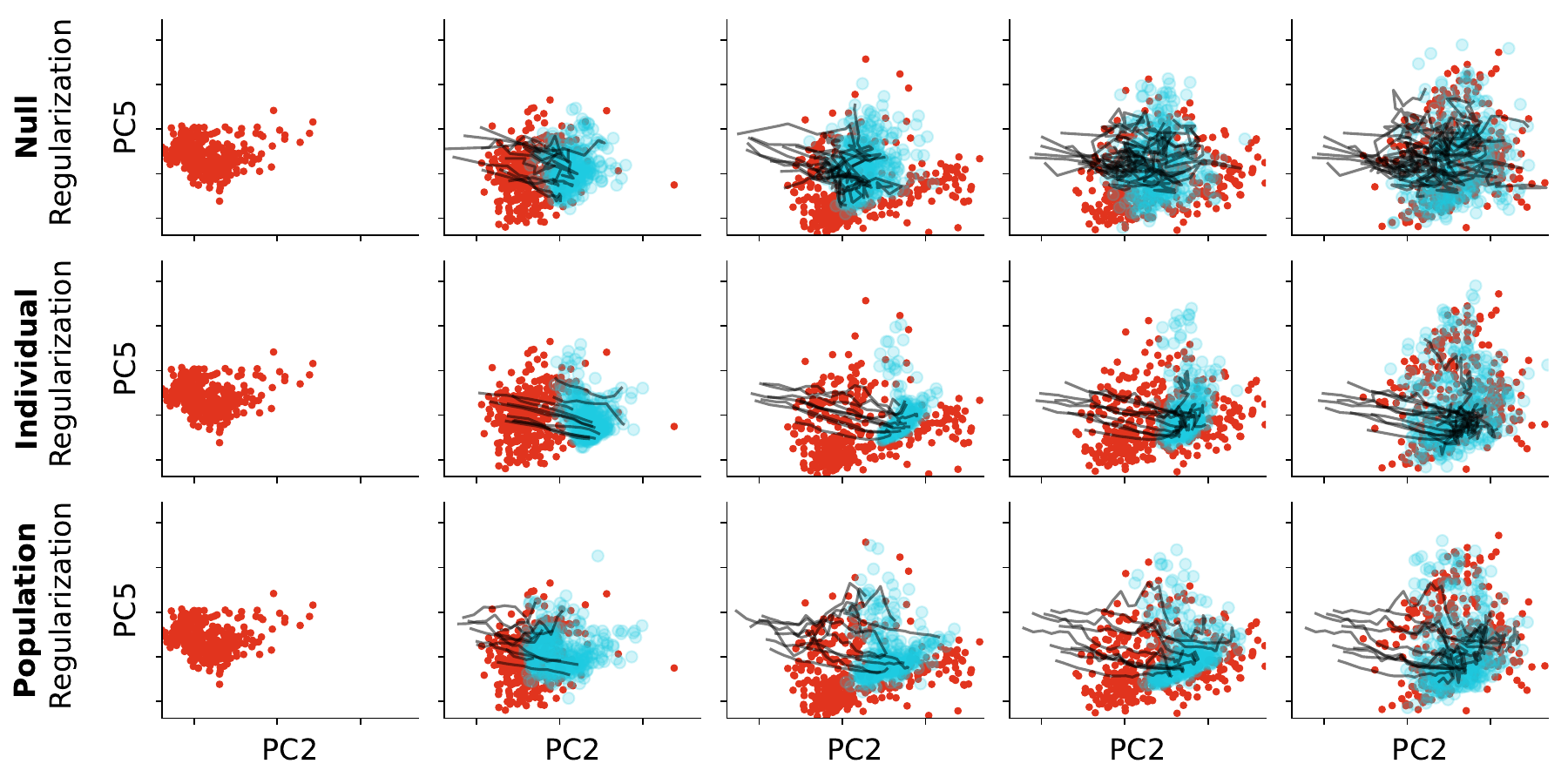}
    \caption{Visualization of global dynamics for principle components 2 and 5.}
\end{figure}

\begin{figure}[!htb] 
    \centering 
    \includegraphics[width=1\linewidth]{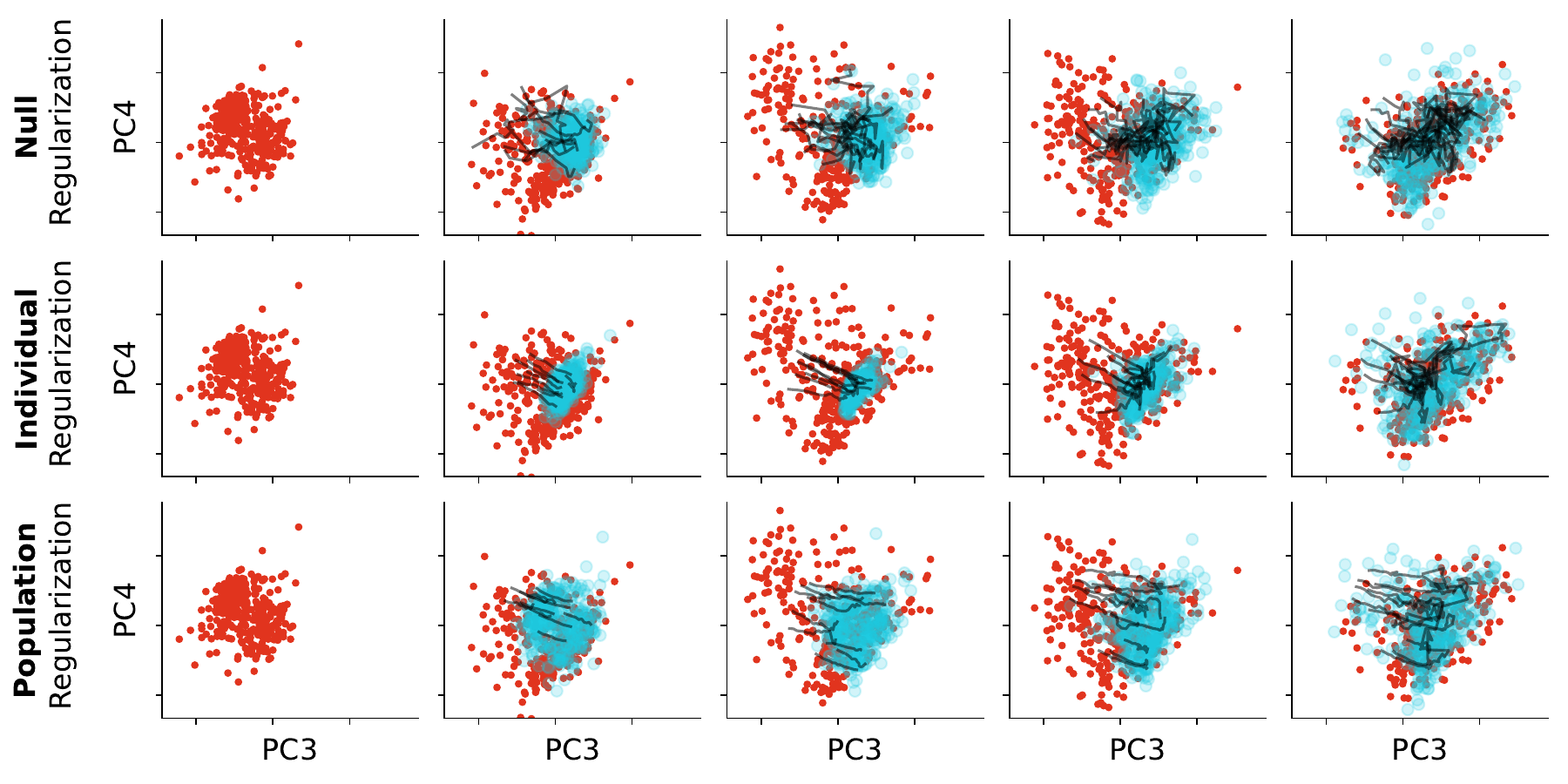}
    \caption{Visualization of global dynamics for principle components 3 and 4.}
\end{figure}

\begin{figure}[!htb] 
    \centering 
    \includegraphics[width=1\linewidth]{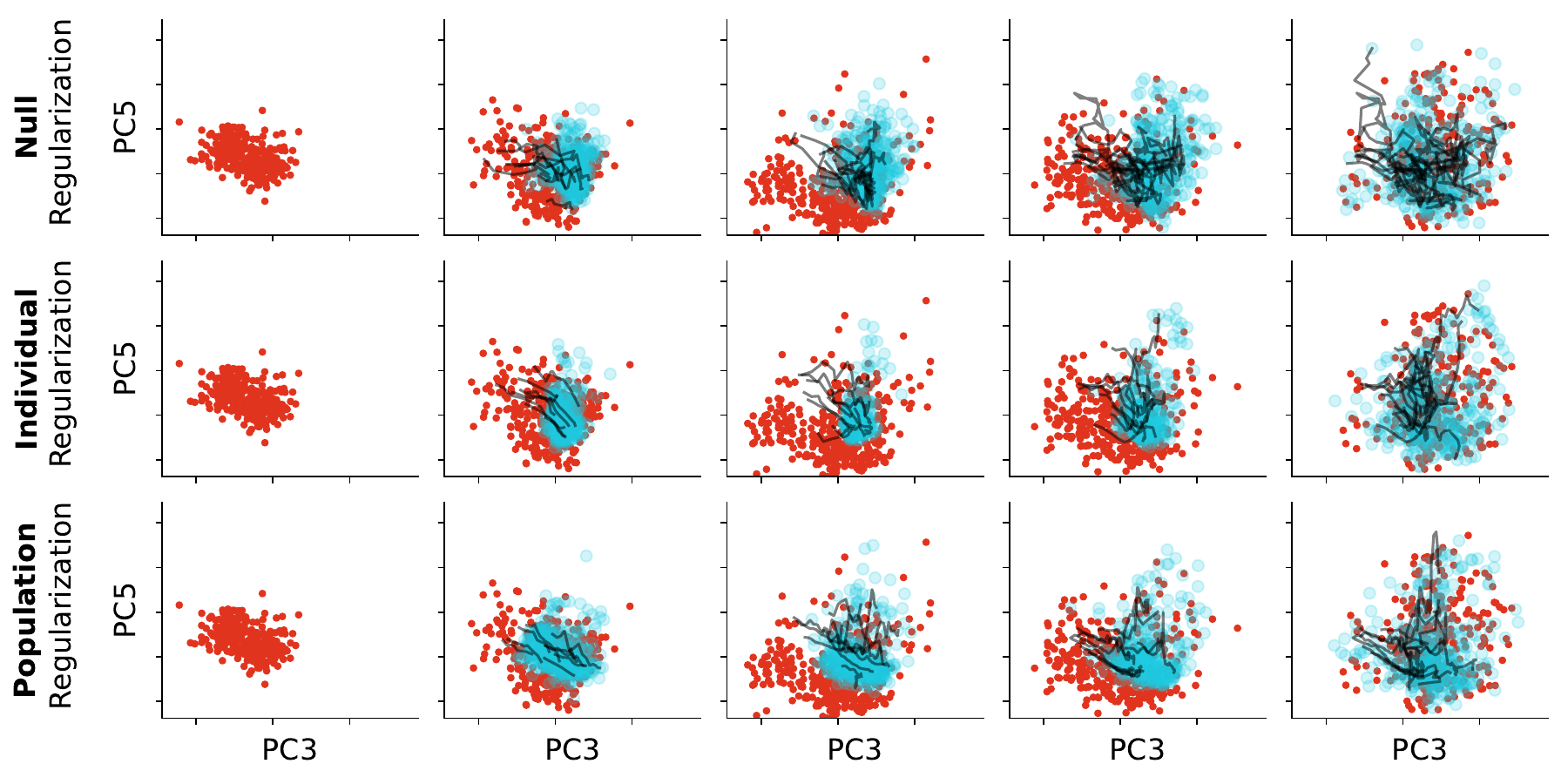}
    \caption{Visualization of global dynamics for principle components 3 and 5.}
\end{figure}

\begin{figure}[!htb] 
    \centering 
    \includegraphics[width=1\linewidth]{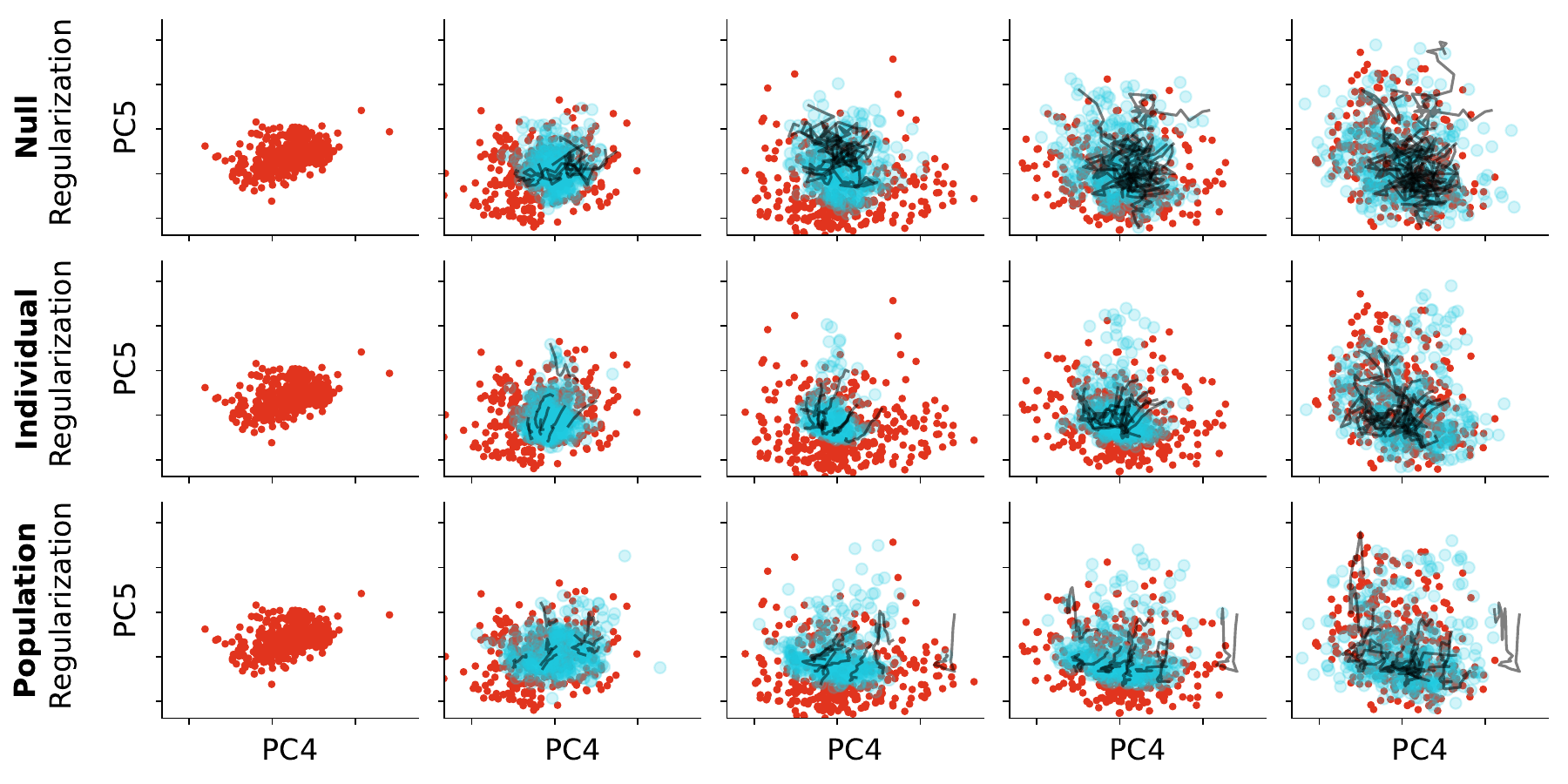}
    \caption{Visualization of global dynamics for principle components 4 and 5.}
\end{figure}

\begin{table}[!htb]
\begin{center}
\caption{Evaluation in the unconditional generation scenario of developmental modeling of embryonic stem cells.}
\label{tab:ablation_studies}
\resizebox{1\textwidth}{!}{
\begin{tabular}{c | c c c | c c c c c c}
    \toprule
    \multirow{2}{*}{Methods} & \multicolumn{3}{c|}{All-Step Prediction} & \multicolumn{3}{c}{One-Step Prediction}\\
    & $t_1$ & $t_2$ & $t_3$ & $t_1$ & $t_2$ & $t_3$ \\
    \midrule
    $\alpha_{\text{corr}, 0} = 0, \alpha_{\text{corr}, 1} = 0, \alpha_{\text{corr}, 2} = 0$ & 1.499$\pm$0.005 & 1.945$\pm$0.006 & 1.619$\pm$0.016 & 1.498$\pm$0.005 & 1.418$\pm$0.009 & 0.966$\pm$0.016 \\
    \midrule
    $\alpha_{\text{corr}, 0} > 0, \alpha_{\text{corr}, 1} = 0, \alpha_{\text{corr}, 2} = 0$ & 1.468$\pm$0.005 & 1.908$\pm$0.007 & 1.586$\pm$0.015 & 1.467$\pm$0.004 & 1.416$\pm$0.009 & 0.957$\pm$0.016 \\
    $\alpha_{\text{corr}, 0} = 0, \alpha_{\text{corr}, 1} > 0, \alpha_{\text{corr}, 2} = 0$ & 1.035$\pm$0.005 & 1.557$\pm$0.012 & 1.523$\pm$0.021 & 1.034$\pm$0.005 & 1.164$\pm$0.011 & 0.865$\pm$0.014 \\
    $\alpha_{\text{corr}, 0} = 0, \alpha_{\text{corr}, 1} = 0, \alpha_{\text{corr}, 2} > 0$ & 0.946$\pm$0.004 & 1.503$\pm$0.007 & 1.440$\pm$0.015 & 0.946$\pm$0.004 & 1.205$\pm$0.008 & 0.917$\pm$0.013 \\
    \bottomrule
\end{tabular}}
\end{center}
\end{table}

\end{document}